\documentclass[11pt]{article}
\usepackage{acl}

\usepackage{times}
\usepackage{latexsym}

\usepackage[T1]{fontenc}
\usepackage[utf8]{inputenc}

\usepackage{microtype}
\usepackage{inconsolata}
\usepackage{graphicx}
\usepackage{subcaption}
\usepackage[strings]{underscore}

\usepackage{amsmath}
\usepackage{amssymb}
\usepackage{amsfonts}
\usepackage{mathtools}
\usepackage{amsthm}
\usepackage{multirow}

\usepackage{booktabs}
\usepackage{tabularx}
\usepackage{float}
\usepackage{placeins}
\usepackage[table]{xcolor}
\usepackage{enumitem}
\usepackage{ulem}
\usepackage{xspace}
\usepackage[capitalize,noabbrev]{cleveref}

\usepackage[textsize=tiny]{todonotes}

\definecolor{resultgreen}{RGB}{34,139,34}
\definecolor{resultred}{RGB}{178,34,34}
\definecolor{resultgray}{RGB}{120,120,120}
\definecolor{baselinegray}{RGB}{240,240,240}
\newcommand{\score}[2]{%
  \makebox[4.2em][l]{#1\raisebox{-0.75ex}{\hspace{0.25em}{\scriptsize #2}}}%
}

\newcommand{\ie}{\textit{i.e.}\xspace}
\newcommand{\eg}{\textit{e.g.}\xspace}
\newcommand{\method}{\textsc{POISE}\xspace}

\title{From AI Assistant to AI Scientist: Autonomous Discovery of LLM-RL Algorithms with LLM Agents}

\author{
    Sirui Xia$^{1*}$, Yikai Zhang$^{1}$\thanks{~The first two authors contributed equally.}, Aili Chen$^{1}$, Siye Wu$^{1}$, Siyu Yuan$^{2}$, Yanghua Xiao$^{1}$\thanks{~Corresponding authors.} \\
    $^1$Shanghai Key Laboratory of Data Science, School of Computer Science, Fudan University \\
    $^2$School of Data Science, Fudan University \quad \\
    \texttt{\{srxia24, ykzhang22\}@m.fudan.edu.cn,}
    \texttt{shawyh@fudan.edu.cn} \\  
}

\begin{document}
\maketitle


\begin{abstract}
Discovering improved policy optimization algorithms for language models remains a costly manual process requiring repeated mechanism-level modification and validation.
Unlike simple combinatorial code search, this problem requires searching over algorithmic mechanisms tightly coupled with training dynamics while reusing empirical evidence across iterations.
We propose \method{}, a closed-loop framework for automated discovery of policy optimization algorithms for language models.
\method{} maintains a structured, genealogically linked archive linking proposals, executable implementations, standardized evaluations, and natural-language reflections to support evidence-driven iteration.
In mathematical reasoning experiments starting from GRPO, \method{} evaluates 64 candidate algorithms and discovers improved mechanisms, including analytic-variance scaling and validity masking.
The best variant improves weighted Overall from 47.8 to 52.5 (+4.6) and increases AIME25 \texttt{pass@32} from 26.7\% to 43.3\%, demonstrating the feasibility of automated policy optimization discovery while supporting interpretable design principles.
\end{abstract}


\section{Introduction}
\label{sec:introduction}

Reinforcement Learning (RL) has become an important post-training paradigm for Large Language Models (LLMs)~\citep{ouyang2022training, shao2024deepseekmath}.
However, discovering improved policy optimization algorithms for language models remains a costly manual process.
Designing an effective policy optimization algorithm involves more than hyperparameter tuning: researchers repeatedly modify and validate core mechanisms, including loss design, advantage estimation, and regularization.
Each new hypothesis requires time-consuming, resource-intensive training and evaluation, burdening researchers and demanding substantial computational resources.
LLMs have shown advanced reasoning and planning capabilities~\citep{yang2025qwen3, liu2025deepseek, erdoganplan}, and recent studies have begun developing LLM-driven systems for automated research~\citep{lu2024ai, schmidgall2025agent, schmidgall2025agentrxiv}.
This raises a question: can LLMs discover new policy optimization algorithms for language models?

Although automated research has made remarkable progress in tasks such as machine-learning research automation~\citep{liu2025ml}, program optimization~\citep{fawzi2023funsearch}, and model architecture discovery~\citep{liu2025alphago}, discovering new policy optimization algorithms for language models still presents unique challenges.
Unlike tasks framed as combinatorial searches over discrete program components, discovering policy optimization algorithms for language models requires searching over mechanisms tightly coupled with training dynamics.
For example, small design choices, such as introducing a clipping objective as in PPO~\citep{schulman2017proximal}, adding a KL penalty, or modifying advantage normalization, can substantially alter optimization behavior and training stability.
Moreover, discovering a new algorithm requires more than proposing an implementation: it also requires identifying which mechanisms contribute to performance gains, which fail under empirical evaluation, and how accumulated evidence should inform the next testable hypothesis.
This calls for a search process that systematically converts empirical feedback into reusable evidence and carries that evidence across iterations.

To address this challenge, we propose \method{} (\textbf{P}olicy \textbf{O}ptimization through \textbf{I}terative \textbf{S}earch and \textbf{E}vidence), a closed-loop framework for the automated discovery of policy optimization algorithms for language models.
Within \method{}, algorithm discovery is carried out as Epistemic Evolutionary Search over an algorithm space with a reflection-augmented evolutionary solver.
The framework maintains a structured archive with genealogical links that associates each algorithm proposal with its executable implementation, standardized evaluation outcomes, and natural-language reflections, allowing later proposals to build on accumulated evidence instead of isolated trial-and-error.
In mathematical reasoning experiments starting from a GRPO baseline~\citep{shao2024deepseekmath}, \method{} evaluates 64 algorithms and discovers improved policy optimization mechanisms, including analytic-variance scaling and validity masking.
The best variant improves weighted Overall from 47.8 to 52.5 (+4.6) and increases AIME25 \texttt{pass@32} from 26.7\% to 43.3\%, while the resulting lineages yield interpretable design principles.

Our main contributions are as follows:
\begin{itemize}[leftmargin=*, nosep]
    \item \textbf{Closed-Loop Discovery Framework:} We propose \method{}, a closed-loop framework for automated discovery of policy optimization algorithms for language models.
    It combines reflection augmentation with a structured, genealogically linked archive supporting evidence-driven iteration from proposal generation to implementation, verification, evaluation, and archive update.
    \item \textbf{Empirical Validation of Automated Algorithm Discovery:} Starting from GRPO, \method{} evaluates 64 candidate algorithms and discovers mechanisms that yield substantial gains on mathematical reasoning benchmarks.
    These results demonstrate the feasibility of automated policy optimization discovery for language models.
    \item \textbf{Interpretable Design Principles:} Beyond discovering stronger algorithms, our analysis extracts design principles from lineage evolution, including signal decoupling, conditional normalization, and correctness-first efficiency shaping.
    These principles clarify how and when such mechanisms help in sparse-reward language model policy optimization.
\end{itemize}

\section{Related Work}
\label{sec:related_work}

\paragraph{LLM-driven Scientific Discovery} leverages the knowledge, reasoning, and execution capabilities of Large Language Models (LLMs) to automate parts of the research process.
This paradigm has evolved from basic experimental automation~\citep{boiko2023autonomous} to complex algorithm design~\citep{fawzi2023funsearch,shihierarchically}.
More recently, ``AI Scientist'' agents have emerged, aiming for end-to-end pipelines that encompass literature review, coding, and experimentation~\citep{lu2024ai,yamada2025ai,schmidgall2025agent,gottweis2025towards,mitchener2025kosmos}.
Progress has also been made in quality control for idea generation~\citep{sican,li2024learning,keya2025sci}.
Furthermore, LLMs have demonstrated effective exploration within well-defined search spaces across domains like program optimization~\citep{novikov2025alphaevolve}, theorem proving~\citep{trinh2024solving,chervonyi2025gold}, architecture search~\citep{liu2025alphago,cheng2025language}, and automated machine learning engineering~\citep{chanmle,jiang2025aide,liu2025ml}.
However, extending automated discovery to RL algorithm design for language model policy optimization is more challenging because the search space is compositional, feedback is delayed and stochastic, and verification requires expensive distributed training runs.
To address this, we propose \method{}, an iterative evolutionary system guiding LLMs to design, verify, and refine policy-optimization algorithms for language models.

\paragraph{Policy Optimization for Large Language Models} aims to enhance reasoning capabilities by training LLMs with algorithms like PPO~\citep{schulman2017proximal} and its computationally efficient variant GRPO~\citep{shao2024deepseekmath}.
Recent studies have enhanced these baselines by refining specific algorithmic components, such as decoupled clipping in DAPO~\citep{yu2025dapo}, soft adaptive gating in SAPO~\citep{gao2025soft}, asymmetric weighting in ASPO~\citep{wang2025aspo}, and geometric mean aggregation in GMPO~\citep{zhao2025geometric}.
While these studies demonstrate the value of targeted component-level refinements, exploring the combinatorial design space of such mechanisms still largely relies on manual iteration.
Unlike prior work, which typically relies on manually refining individual algorithmic components, our method introduces an LLM-driven evolutionary system to autonomously discover and compose algorithmic improvements under standardized settings.


\section{Method}
\label{sec:method}

\begin{figure*}[t!]
  \centering
  \includegraphics[width=0.98\textwidth]{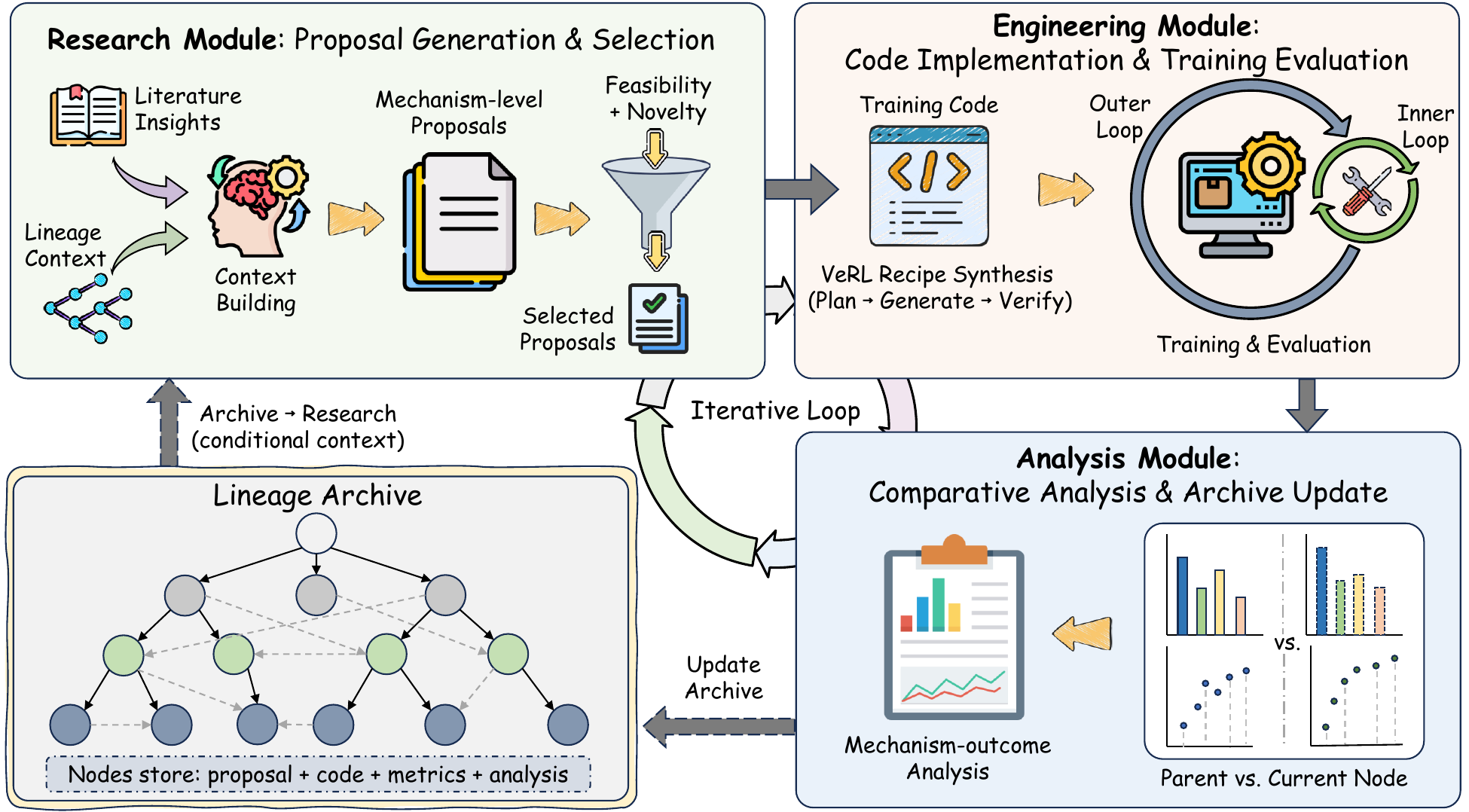}
  \caption{Overview of \method{}.
  Phase I (Proposal Generation) generates candidate algorithms from historical evidence and prior knowledge;
  Phase II (Implementation, Verification, and Evaluation) implements proposals under shared interfaces, verifies fidelity to the intended algorithm, and evaluates them with a standardized protocol;
  Phase III (Reflective Analysis and Archive Update) interprets the results and updates the archive.}
  \label{fig:method}
\end{figure*}

\subsection{Problem Modeling and Methodology}
\label{sec:method:modeling}

We formulate automated discovery of policy optimization algorithms for language models as search over a space of feasible reinforcement learning algorithms.
Let \(\mathcal{Z}\) denote the space of feasible algorithms supported by our framework, where each algorithm \(z \in \mathcal{Z}\) specifies a complete training procedure through the composition of algorithmic components (\eg, loss functions, advantage estimators) and their associated hyperparameters.

For any algorithm \(z \in \mathcal{Z}\), a standardized evaluation pipeline \(\mathcal{O}(z)\) executes the training-and-evaluation workflow and returns a training trajectory \(\tau\) and a vector of evaluation metrics \(\mathbf{y} \in \mathbb{R}^K\), \ie\ \((\tau, \mathbf{y}) = \mathcal{O}(z)\).
The trajectory \(\tau\) captures training dynamics, such as learning curves, gradient norms, and stability signals, while \(\mathbf{y}\) quantifies performance across \(K\) evaluation metrics.

Our objective is to discover an algorithm \(z^*\) that maximizes the expected scalar utility of the resulting metric vector:
\begin{equation}
\label{eq:objective}
  z^* = \mathop{\arg\max}_{z \in \mathcal{Z}} \;
        \mathbb{E}
        \left[ U\big(\mathbf{y}(z)\big) \right],
\end{equation}
%
where the expectation is taken over the randomness of training and evaluation, and \(U: \mathbb{R}^K \to \mathbb{R}\) maps the \(K\) evaluation metrics to a scalar utility.

Within \method{}, algorithm discovery is implemented as \textbf{Epistemic Evolutionary Search over an algorithm space}, where \emph{epistemic} underscores that search must reason under uncertainty about which design choices and lineages remain promising given accumulated archive evidence.
As in Figure~\ref{fig:method}, \method{} organizes this search into three phases: proposal generation (Phase I), implementation, verification, and evaluation (Phase II), and reflective analysis with archive update (Phase III).
A reflection-augmented evolutionary solver maintains a \textbf{structured archive with genealogical links}, denoted \(\mathcal{M}_t = \{e_1, \dots, e_t\}\), across phases.
Each entry \(e_i = (z_i, \rho_i, \tau_i, \mathbf{y}_i, r_i)\) records the algorithm proposal, its executable implementation \(\rho_i\), resulting empirical evidence as training trajectories \(\tau_i\) and evaluation metrics \(\mathbf{y}_i\), and natural-language reflection \(r_i\).
In Phase II, the evaluation pipeline \(\mathcal{O}\) implements a proposal under shared interfaces, applies verification and consistency checks, and evaluates it under a standardized protocol. This evidence is then consumed by the reflective analysis and archive update operators in Phase III.
Around this evaluation pipeline, the closed loop is coordinated by three operators:

\begin{enumerate}[noitemsep,left=0pt]
    \item \textbf{Proposal Operator ($\mathcal{P}$):}
    Given a selected parent entry \(e_p \in \mathcal{M}_t\), it proposes improved candidates in the algorithm space conditioned on historical evidence, reflection, and external literature priors \(\mathcal{L}\):
    \[
        \{z_{t+1}^{(n)}\}_{n=1}^N \sim \mathcal{P}_{\text{LLM}}\left(\cdot \mid z_p, r_p, \mathcal{M}_t, \mathcal{L}\right).
    \]

    \item \textbf{Reflective Analysis Operator ($\mathcal{A}$):}
    After evaluation via \((\tau_{t+1}, \mathbf{y}_{t+1}) = \mathcal{O}(z_{t+1})\), this operator converts training dynamics and metric outcomes into a mechanism-level diagnosis:
    \[
        r_{t+1} = \mathcal{A}_{\text{LLM}}(z_{t+1}, \tau_{t+1}, \mathbf{y}_{t+1} \mid \mathcal{M}_t).
    \]

    \item \textbf{Archive Update Operator ($\mathcal{U}$):}
    It updates the archive and lineage structure, and selects promising anchors for subsequent evolution:
    \[
        \mathcal{M}_{t+1} = \mathcal{U}\big(\mathcal{M}_t, z_{t+1}, \rho_{t+1}, \tau_{t+1}, \mathbf{y}_{t+1}, r_{t+1}\big).
    \]
\end{enumerate}

This formulation organizes algorithm search as an evidence-driven closed loop.

\subsection{Phase I: Proposal Generation via Epistemic Evolutionary Search}
\label{sec:method:phase1}

The proposal operator \(\mathcal{P}\) generates high-quality candidates conditioned on archive evidence, reflection, and literature priors \(\mathcal{L}\).
It prioritizes promising parents, constructs proposal context, and selects candidates for subsequent evaluation.

\paragraph{Lineage Prioritization.}
To identify evolutionary anchors, we construct an acquisition function \(s(d)\) that balances strong existing lineages against epistemic uncertainty:
Parent selection should therefore favor not only strong nodes, but also nodes likely to yield stronger descendants.
\[
\begin{split}
s(d) &= w_{\text{pareto}} \cdot \mathcal{U}_{\text{pareto}} + w_{\text{perf}} \cdot \mathcal{U}_{\text{perf}} \\
&\quad + w_{\text{div}} \cdot \mathcal{U}_{\text{div}} + w_{\text{bayes}} \cdot \alpha_{\text{GP}}(d).
\end{split}
\]
Here, \(\mathcal{U}_{\text{pareto}}\), \(\mathcal{U}_{\text{perf}}\), and \(\mathcal{U}_{\text{div}}\) denote Pareto-frontier strength, normalized performance, and diversity signals, respectively, while the weights \(\mathbf{w}\) control their balance.
Crucially, \(\alpha_{\text{GP}}(d)\) provides an estimate of a node's descendant potential, namely whether it is likely to yield stronger descendants rather than merely scoring well currently.
To avoid favoring immediate performance too myopically, we train this GP signal on a \textbf{Discounted Top-K Descendant Gain} target, where \(U(\cdot)\) denotes the scalar node utility from the evaluation metrics, \(c_k\) ranges over the top-K descendants of \(d\), and \(\beta\) sets the gain scale:
\[
\begin{split}
y_d &= \mathbb{E}_{k \in \text{top-K}} \big[ \gamma^{\text{depth}(c_k, d)} \\
&\quad \cdot \tanh\big((U(c_k) - U(d))/\beta\big) \big],
\end{split}
\]
We estimate \(\alpha_{\text{GP}}\) with a GP-based UCB term over archived node features to preserve sensitivity to epistemic uncertainty.

\paragraph{Context Construction from the Archive and Literature.}
For a selected parent \(p\), we construct a hybrid proposal context from archive evidence, lineage structure, and external priors.
To provide complementary reference points, we assemble a tiered reference set: (i) Pareto-frontier references; (ii) complementary references chosen for structural contrast relative to the parent; and (iii) exploratory references sampled outside the direct lineage.
We further incorporate literature priors from \(\mathcal{L}\) when available.
Together, these archive cases and literature priors shape proposal generation.

\paragraph{Population Evolution \& Selection.}
From this context, the proposal operator identifies promising improvement directions and produces algorithm proposals.
We adopt a \textbf{Population-Based Generation} strategy to sample \(N\) parallel candidates spanning diverse local directions.
A screening stage filters and ranks candidates for logical consistency, implementation feasibility, novelty, and expected improvement relative to the parent.
The highest-ranked candidate \(z_{t+1}\) is selected for implementation, verification, and evaluation.

\begin{table*}[!t]
  \centering
  \footnotesize
  \renewcommand{\arraystretch}{0.9}
  \setlength{\tabcolsep}{2.8pt}
  \providecommand{\resulttablewidth}{0.98\textwidth}
  \begin{tabular*}{\resulttablewidth}{@{\extracolsep{\fill}} llllllll}
    \toprule
    \textbf{Algorithm} &
    \shortstack{\textbf{AIME24}\\{\small\texttt{pass@32}}} &
    \shortstack{\textbf{AIME25}\\{\small\texttt{pass@32}}} &
    \shortstack{\textbf{AMC}\\{\small\texttt{pass@32}}} &
    \shortstack{\textbf{MATH}\\{\small\texttt{acc@1}}} &
    \shortstack{\textbf{Minerva}\\{\small\texttt{acc@1}}} &
    \shortstack{\textbf{Olympiad}\\{\small\texttt{acc@1}}} &
    \textbf{Overall} \\
    \midrule
    GRPO & 50.0 & 26.7 & 84.3 & 72.4 & 24.6 & 35.4 & 47.8 \\
    \specialrule{0.6pt}{0.4ex}{0.0ex}
    \rowcolor{gray!18}
    \multicolumn{8}{c}{\rule[-0.6ex]{0pt}{2.8ex}\textit{Core Signal Modeling}} \\
    \addlinespace[0.35ex]
    VM-AV-GRPO & \score{53.3}{\textcolor{resultgreen}{+3.3}} & \score{43.3}{\textcolor{resultgreen}{+16.7}} & \score{88.0}{\textcolor{resultgreen}{+3.6}} & \score{73.6}{\textcolor{resultgreen}{+1.2}} & \score{24.3}{\textcolor{resultred}{-0.4}} & \score{35.3}{\textcolor{resultred}{-0.1}} & \score{52.5}{\textcolor{resultgreen}{+4.6}} \\
    AV-GRPO & \score{56.7}{\textcolor{resultgreen}{+6.7}} & \score{36.7}{\textcolor{resultgreen}{+10.0}} & \score{81.9}{\textcolor{resultred}{-2.4}} & \score{71.0}{\textcolor{resultred}{-1.4}} & \score{26.8}{\textcolor{resultgreen}{+2.2}} & \score{35.3}{\textcolor{resultred}{-0.1}} & \score{50.9}{\textcolor{resultgreen}{+3.1}} \\
    \specialrule{0.6pt}{0.4ex}{0.0ex}
    \rowcolor{gray!18}
    \multicolumn{8}{c}{\rule[-0.6ex]{0pt}{2.8ex}\textit{High-Performing Specializations}} \\
    \addlinespace[0.35ex]
    MSA-GRPO & \score{50.0}{\textcolor{resultgray}{+0.0}} & \score{43.3}{\textcolor{resultgreen}{+16.7}} & \score{84.3}{\textcolor{resultgray}{+0.0}} & \score{69.4}{\textcolor{resultred}{-3.0}} & \score{24.3}{\textcolor{resultred}{-0.4}} & \score{35.3}{\textcolor{resultred}{-0.1}} & \score{50.7}{\textcolor{resultgreen}{+2.8}} \\
    SVE-LNA-GRPO & \score{50.0}{\textcolor{resultgray}{+0.0}} & \score{30.0}{\textcolor{resultgreen}{+3.3}} & \score{89.2}{\textcolor{resultgreen}{+4.8}} & \score{73.0}{\textcolor{resultgreen}{+0.6}} & \score{23.2}{\textcolor{resultred}{-1.5}} & \score{32.7}{\textcolor{resultred}{-2.7}} & \score{48.7}{\textcolor{resultgreen}{+0.9}} \\
    \specialrule{0.6pt}{0.4ex}{0.0ex}
    \rowcolor{gray!18}
    \multicolumn{8}{c}{\rule[-0.6ex]{0pt}{2.8ex}\textit{Failure-side and Routing Variants}} \\
    \addlinespace[0.35ex]
    FA-GRPO & \score{53.3}{\textcolor{resultgreen}{+3.3}} & \score{36.7}{\textcolor{resultgreen}{+10.0}} & \score{83.1}{\textcolor{resultred}{-1.2}} & \score{72.2}{\textcolor{resultred}{-0.2}} & \score{22.4}{\textcolor{resultred}{-2.2}} & \score{35.0}{\textcolor{resultred}{-0.4}} & \score{49.9}{\textcolor{resultgreen}{+2.1}} \\
    PR-GRPO & \score{53.3}{\textcolor{resultgreen}{+3.3}} & \score{36.7}{\textcolor{resultgreen}{+10.0}} & \score{80.7}{\textcolor{resultred}{-3.6}} & \score{71.2}{\textcolor{resultred}{-1.2}} & \score{23.9}{\textcolor{resultred}{-0.7}} & \score{33.2}{\textcolor{resultred}{-2.2}} & \score{49.4}{\textcolor{resultgreen}{+1.5}} \\
    CDA-GRPO & \score{50.0}{\textcolor{resultgray}{+0.0}} & \score{30.0}{\textcolor{resultgreen}{+3.3}} & \score{85.5}{\textcolor{resultgreen}{+1.2}} & \score{72.6}{\textcolor{resultgreen}{+0.2}} & \score{25.0}{\textcolor{resultgreen}{+0.4}} & \score{36.9}{\textcolor{resultgreen}{+1.5}} & \score{49.0}{\textcolor{resultgreen}{+1.2}} \\
    \bottomrule
  \end{tabular*}
  \vspace{0.5ex}
  \caption{Held-out reasoning evaluation results (percentage points). We report a compact set of strong and mechanism-representative variants in the main text; the full results table (all evaluated variants) is provided in the Appendix. Small colored numbers denote \(\Delta\) vs.\ GRPO (green: improvement; red: regression). ``Overall'' is the fixed weighted average used by the evolutionary selector (weights in \cref{sec:exp:setup}). ``Olympiad'' denotes OlympiadBench.}
  \label{tab:experiment:main_results}
\end{table*}

\subsection{Phase II: Implementation, Verification, and Evaluation}
\label{sec:method:phase2}

The evaluation pipeline \(\mathcal{O}\) maps abstract proposals to reliable, comparable empirical evidence through implementation, verification, training, and standardized evaluation.
This stage emphasizes faithful implementation, consistency checks, and standardized evaluation so observed differences can be attributed as much as possible to algorithmic changes rather than implementation mismatches.

\paragraph{Implementation and Verification under Shared Interfaces.}
Each proposal is implemented under a common training interface so algorithmic changes remain executable and directly comparable.
Given a proposal \(z_{t+1}\), the framework instantiates it as implementation \(\rho_{t+1}\) by modifying core optimization components such as advantage computation, objective design, or regularization.
During implementation, the framework uses interface constraints and consistency checks to align the implementation with the intended algorithm.
Before full evaluation, if verification or quick validation reveals clear mismatches with design intent or implementation faults, the candidate is revised within bounded correction loops so the resulting evidence remains reliable and comparable.

\paragraph{Standardized Evaluation Protocol.}
For fair comparison, candidates are evaluated under a unified protocol with aligned training budgets, backend settings, and metric definitions.
The framework aggregates training trajectories \(\tau_{t+1}\) and executes standardized tests to compute the evaluation metrics \(\mathbf{y}_{t+1}\).
Thus, archived nodes remain comparable and suitable for later attribution and selection.

\subsection{Phase III: Reflective Analysis \& Archive Update}
\label{sec:method:phase3}

\paragraph{Reflective Analysis.}
After evaluation, the Reflective Analysis Operator \(\mathcal{A}\) converts quantitative signals into mechanism-level diagnoses.
It analyzes the algorithmic change \(\Delta z\), the training trajectories \(\tau_{t+1}\), and the evaluation metrics \(\mathbf{y}_{t+1}\), and contrasts them with the parent to identify likely contributing effects.
For example, it can relate KL spikes to numerical instability in advantage estimation, or connect gains to a particular signal-routing mechanism.
Outcomes thus become reusable evidence rather than isolated score changes.

\paragraph{Epistemic Accumulation and Archive Update.}
The archive stores not only code and scores but also which design choices help, fail, or depend on context.
As entries are integrated into \(\mathcal{M}\), the framework updates its understanding of which design choices are effective, which are fragile, and when they remain valid.
Reflections on strong nodes reinforce useful patterns; those on weak nodes delineate failure boundaries.
Over time, the archive becomes a practical reference for proposals, and the regularities from it feed back into proposal generation as inductive bias.

\section{Experiments}
\label{sec:experiments}

\begin{figure*}[t]
  \centering
  \includegraphics[width=\linewidth]{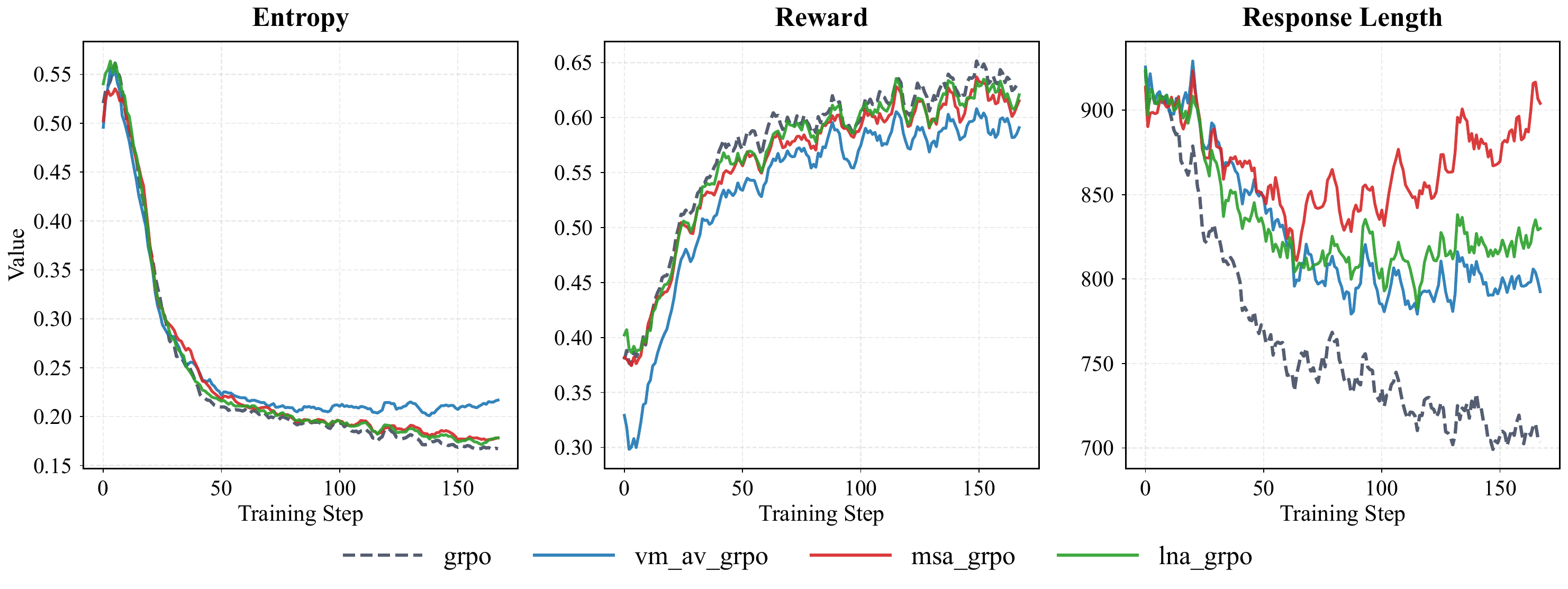}
  \caption{Training dynamics in the main experiment for GRPO and representative evolved variants.
  The plots compare entropy, reward, and response length over training.}
  \label{fig:training_dynamics_main}
\end{figure*}

\subsection{Experimental Setup}
\label{sec:exp:setup}

\paragraph{Training Setup.}
We implement candidate algorithms in a shared VERL training pipeline~\citep{sheng2024hybridflow}, running on its distributed policy optimization backend.
Algorithm evolution starts from the \textbf{GRPO} baseline~\citep{shao2024deepseekmath}, with \textbf{Gemini-3-pro-preview}~\citep{gemini-3-pro} serving as the unified reasoning engine for all framework agents.
Following~\citet{zhao2025geometric}, we use \textbf{Qwen2.5-Math-1.5B}~\citep{yang2024qwen2} as the policy model for rapid iteration.
For training data, we sample a \(5000\)-example subset from MATH Levels 3--5~\citep{hendrycksmath2021}.
Hyperparameters remain fixed across iterations: learning rate \(10^{-6}\), global batch size \(256\), group size \(G=8\), and \(8\) training epochs.
Maximum prompt and response lengths are set to \(1024\) and \(3000\), respectively.
All runs use \(8\times\) 80GB A100 GPUs.

\paragraph{Evaluation Protocol.}
We evaluate on a standardized reasoning suite spanning six datasets: \emph{AIME24}~\citep{AIME24}, \emph{AIME25}~\citep{AIME25}, \emph{AMC}~\citep{AMC23}, \emph{MATH-500}~\citep{hendrycksmath2021}, \emph{Minerva}~\citep{Minerva-Math}, and \emph{OlympiadBench}~\citep{OlympiadBench}.
Inference uses vLLM~\citep{kwon2023efficient} with a maximum generation length of \(4096\).
For AIME24, AIME25, and AMC, we report \(\mathrm{pass}@32\) under temperature \(1.0\).
For the larger datasets MATH-500, Minerva, and OlympiadBench, we report standard \(\mathrm{acc}@1\) using deterministic generation.
To align with our evolutionary objective, we compute a weighted \textbf{Overall} favoring harder benchmarks (\(0.2\) for AIME-series datasets and \(0.15\) for the others).

\subsection{Results}
\label{sec:exp:results}

\paragraph{Main Results.}
In the main evolution run we evaluate \(64\) algorithms: GRPO and \(63\) evolved variants.
\cref{tab:experiment:main_results} highlights representative algorithms, with full results in \cref{sec:appendix:full_results}.
The strongest overall variant is VM-AV-GRPO, achieving an Overall of \(52.5\) and improving over GRPO by \(+4.6\).
Its largest gain appears on AIME25, where performance rises from \(26.7\) to \(43.3\), while AIME24, AMC, and MATH also improve over baseline.
The search also uncovers benchmark-specific variants.
Beyond VM-AV-GRPO, AV-GRPO attains the best AIME24 result (\(56.7\)), MSA-GRPO matches the best AIME25 score (\(43.3\)), and SVE-LNA-GRPO achieves the highest AMC score (\(89.2\)).
These results show evolutionary search improves overall performance and discovers variants with stronger results on specific benchmarks.

\paragraph{Mechanism Families.}
Across the lineage, four recurrent mechanism families emerge.
\textbf{Core Signal Modeling} (\eg, AV-GRPO and VM-AV-GRPO) changes how sparse successes are scaled and compared.
\textbf{Signal Routing and Conditional Normalization} (\eg, PR-GRPO and CDA-GRPO) separates solved, failed, and mixed regimes and reduces interference among correctness, format, and efficiency signals.
\textbf{Failure-side Control and Failure Reshaping} (\eg, FA-GRPO and DFR-GRPO) provides directional learning signals even when all samples in a group are wrong.
\textbf{Stabilization and Constraint Control} (\eg, RF-GRPO and BN-GRPO) dampens update shocks and policy drift.
This last family mainly serves as enabling or precursor control rather than as the strongest final design, while \cref{sec:analysis,app:detailed_analysis} explain how all four families interact across the lineage.

Figure~\ref{fig:training_dynamics_main} compares the training dynamics of GRPO against representative evolved variants.
Relative to the baseline, stronger variants exhibit more stable entropy and reward trajectories, accompanied by differentiated behaviors in controlling response length.
These process-level patterns suggest systematic differences in optimization behavior rather than isolated fluctuations in final scores.

\subsection{Controllability Under Length-Compression Constraints}
\label{sec:exp:controllability}

To test whether natural-language directives can effectively steer evolutionary search, we introduce a \textbf{length-compression constraint}: the system favors algorithms that substantially reduce reasoning output length while preserving overall accuracy.

\begin{figure*}[t]
  \centering
  \includegraphics[width=\linewidth]{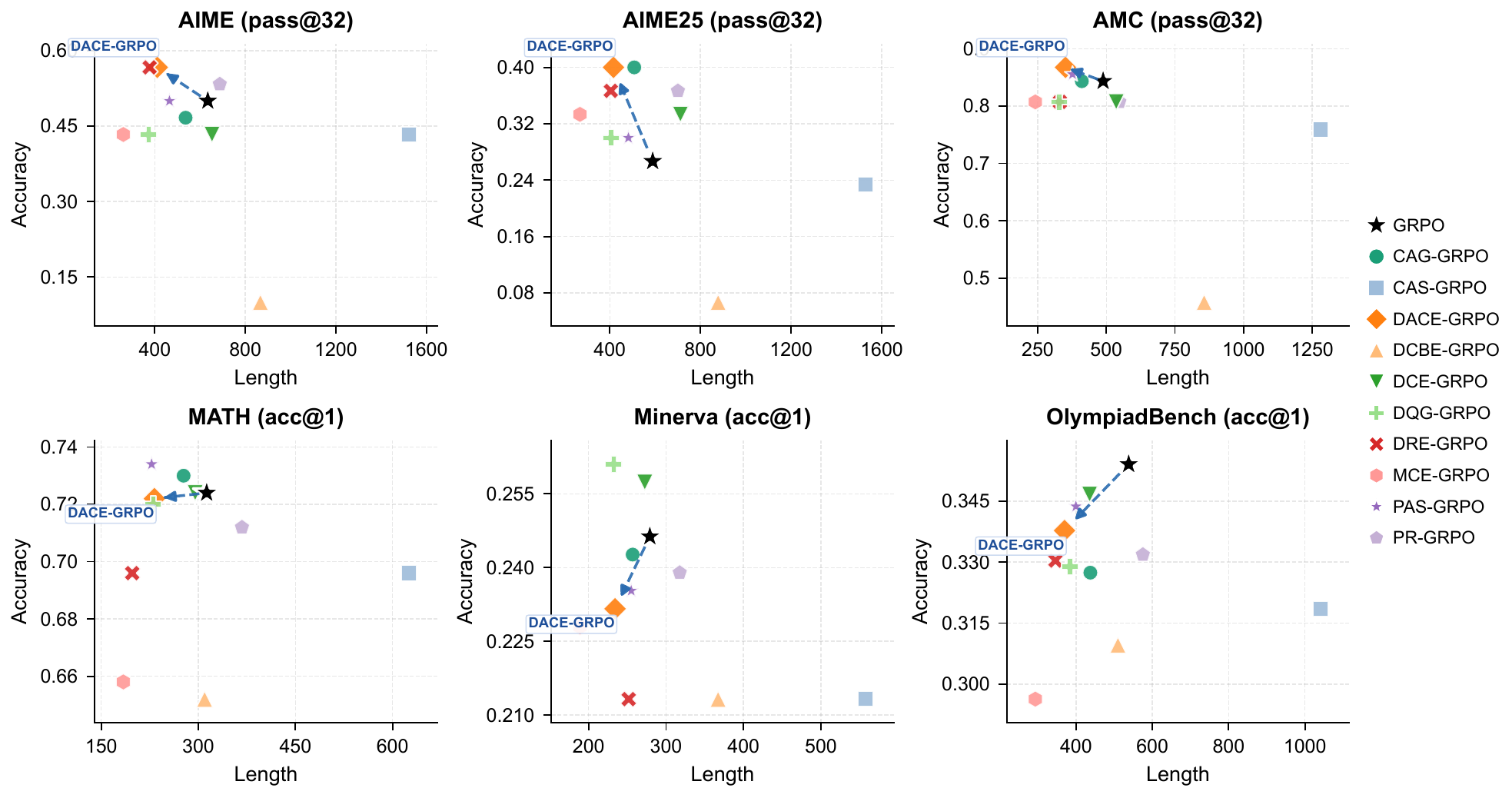}
  \caption{Accuracy-length trade-offs under the length-compression constraint.}
  \label{fig:compression_tradeoff}
\end{figure*}

Under this constraint, \(6\) of the \(10\) evolved variants reduce mean response length.
Mean length is computed as the unweighted mean word count over the six evaluation datasets; \(4\) of these variants (DACE-GRPO, CAG-GRPO, DRE-GRPO, PAS-GRPO) simultaneously achieve shorter responses and higher Overall than GRPO.
The strongest compression results arise from correctness-first optimization combined with conditional efficiency shaping.
For instance, DACE-GRPO gates the length-efficiency term on correct samples by group pass rate, so compression concentrates on high-pass-rate groups while the accuracy signal is strengthened on harder ones.
CAG-GRPO instead reformulates the efficiency term as a penalty-only cost gated by success rate.
Conversely, variants with poorly calibrated efficiency constraints exhibit weaker accuracy-length trade-offs, indicating that compression pressure is more effective when introduced conditionally rather than indiscriminately.
\cref{app:compression_dynamics,tab:compression_compact} summarize the corresponding dynamics and the full quantitative results for all \(11\) nodes in the branch.

Figure~\ref{fig:compression_tradeoff} and Table~\ref{tab:compression_compact} show that DACE-GRPO provides the best balance, improving Overall from \(47.8\) to \(51.7\) (\(+3.9\)) while reducing mean output length from \(473.6\) to \(335.7\) words (\(-29.1\%\)); CAG-GRPO achieves stable compression (length ratio \(0.854\)) with positive Overall gain.
DRE-GRPO and PAS-GRPO show moderate gains but follow the same direction of difficulty- and correctness-conditioned efficiency shaping.
At the dataset level, MATH and AMC retain substantial compression headroom without sacrificing accuracy: for example, DACE-GRPO reaches \(86.7\) on AMC \(\mathrm{pass}@32\) while reducing mean length by \(28.1\%\).
OlympiadBench is more compression-sensitive, where shorter variants generally incur mild accuracy drops, indicating a stronger dependence on longer reasoning chains.
Overall, these results indicate that natural-language directives can meaningfully steer search in a constrained accuracy-length trade-off space, and that correctness-first, conditionally shaped efficiency signals provide a viable route to effective compression.

%

\section{Analysis}
\label{sec:analysis}

\begin{figure}[t]
  \centering
  \includegraphics[width=\linewidth]{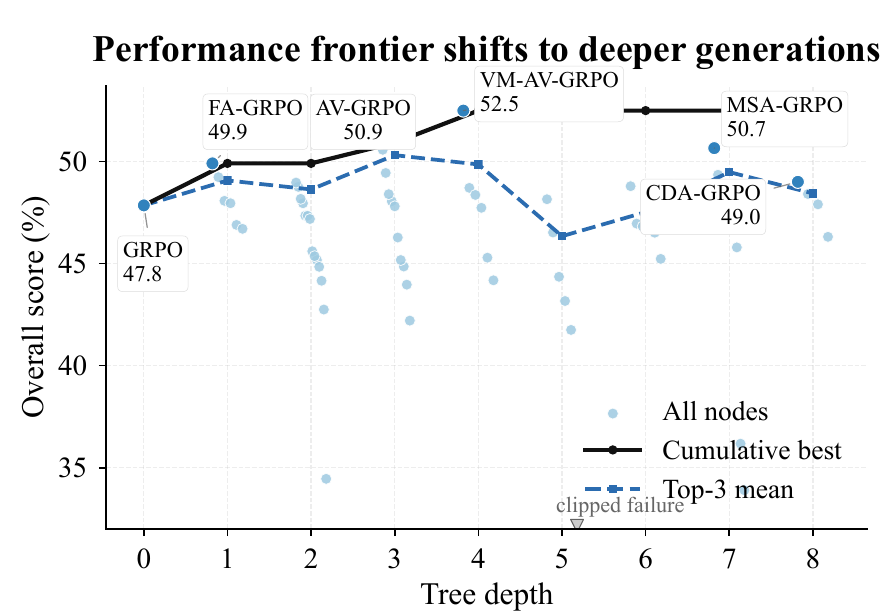}
  \caption{Performance frontier vs.\ tree depth in the base-branch lineage; dots are algorithms.
  Solid: cumulative best Overall; dashed: mean Overall of top-3 at each depth.
  Despite exploratory failures at depth, the frontier improves beyond the root.}
  \label{fig:depth_frontier_base}
\end{figure}

We begin by summarizing the full lineage by tree depth and then examine representative mechanism chains.
Figure~\ref{fig:depth_frontier_base} shows that depth-wise averages are not monotonic, since deeper generations still include exploratory failures.
Nevertheless, the frontier moves beyond the root: the cumulative best Overall increases from \(47.8\) for GRPO to \(49.9\) at depth \(1\), \(50.9\) at depth \(3\), and \(52.5\) at depth \(4\).
Later depths still contain competitive descendants such as MSA-GRPO and CDA-GRPO.
This suggests that the search remains productive even after the best observed node has already appeared.
A complete view by tree depth is provided in \cref{fig:mainrun_local_optimum_base}.

The lineage provides retrospective evidence on parent retention.
Across \(21\) comparable parent-selection rounds (each round selects \(3\) nodes in parallel to expand the next generation), the parent that yields the strongest descendant is not the selected parent with the highest Overall score in \(11\) rounds.
One representative round is as follows (numbers in parentheses denote Overall scores): ZA-GRPO (\(47.3\)), BN-GRPO (\(44.2\)), and Anchor-GRPO (\(45.2\)) are retained together.
In that round, the strongest later descendant arises from BN-GRPO via VM-AV-GRPO (\(52.5\)), above the ZA-GRPO branch peak of \(48.4\).
This pattern is consistent with retaining parents for future improvement rather than filtering by current Overall score.
We provide a fuller round-level account and representative parent-descendant gains in \cref{app:analysis:parent_retention}.

\subsection{Mechanistic Analysis: Core Signal Modeling}
\label{sec:analysis:deep_dive}
A natural question is whether the framework recovers interpretable structure rather than merely tuning hyperparameters.
The lineage from GRPO to VM-AV-GRPO illustrates this in the \textbf{Core Signal Modeling} family, supported by training-dynamics evidence (\cref{fig:training_dynamics_main}), lineage trace (\cref{app:case_study}), and trajectory- and algorithm-level details (\cref{app:analysis:trends,app:algorithms}).
Baseline GRPO underperforms on sparse-reward benchmarks (\eg, AIME25): failure-dominated group statistics can dilute the learning signal from rare successes, while high-variance updates can hasten entropy collapse.

To address this bottleneck, the lineage introduces \textbf{Analytic-variance scaling} (AV-GRPO).
Instead of relying on empirical batch variance, it anchors advantage normalization to the analytic variance of the reward distribution so that rarer successes can yield stronger positive updates.
A subsequent refinement then exposes another failure mode: when "format-correct but reasoning-wrong" samples enter the statistics, they can receive positive advantages.
In response, the lineage introduces \textbf{Validity Masking} (VM-AV-GRPO), which restores a strict gradient hierarchy (invalid $\ll$ valid-wrong $<$ valid-correct) by computing statistics exclusively on the valid subset.
Together, analytic-variance scaling and validity masking are associated with a substantial AIME25 improvement from GRPO to VM-AV-GRPO, while \cref{app:analysis:trends,app:algorithms} trace the intermediate turning points and the corresponding algorithm details behind this refinement.

\subsection{Learning from Negative Evidence}
\label{sec:analysis:epistemic_engine}
Beyond discovering novel algorithms, the framework can also use negative evidence to understand why an algorithm failed.
This is evident in the lineage targeting all-fail stagnation (GRPO $\rightarrow$ FA-GRPO $\rightarrow$ DFR-GRPO), a representative instance of \textbf{Failure-side Control and Failure Reshaping}; \cref{app:analysis:trends,app:algorithms} provide the extended trajectory and algorithm-level formulations.

FA-GRPO first injects gradients into all-fail groups through fixed scalar penalties.
However, post-run analysis of the training dynamics suggests that mixing such fixed penalties with normalized advantages causes a scale mismatch that destabilizes formatting.
Furthermore, applying uniform negative advantages acts as "directionless repulsion," triggering policy drift.
DFR-GRPO then refines this design by separating the signals.
It decouples format and reasoning into independent streams and replaces uniform repulsion with a \textbf{zero-centered signal} derived from token entropy, so that more informative failures are compared against less informative ones rather than being pushed equally.
This yields a more directional gradient even when all candidates are incorrect, alleviating directionless drift, albeit with trade-offs on the hardest benchmarks.
Together with the \textbf{Signal Routing and Conditional Normalization} cases detailed in \cref{app:analysis:trends,app:analysis:lineage_evidence}, this cross-generational refinement highlights three recurring principles: rare successes should be amplified through calibrated signal modeling, flawed hypotheses should be removed through negative evidence, and secondary objectives should be introduced through signal decoupling under controlled conditions.

\section{Conclusion}
\label{sec:conclusion}

In this work, we show that the discovery of policy optimization algorithms for language models can be organized as a structured empirical loop of proposal, implementation, verification, evaluation, and reflection rather than treated solely as manual trial and error.
\method{} discovers stronger policy optimization variants from a GRPO baseline, identifies mechanisms such as analytic-variance scaling and validity masking, and further shows through the length-compression setting that natural-language directives can steer the search toward a specified region of the accuracy-length trade-off space.
Taken together, the resulting lineages reveal recurring principles such as signal decoupling, conditional normalization, and correctness-first efficiency shaping, suggesting that part of algorithm design can be studied as an evidence-driven iterative process that complements direct manual design.

\section*{Limitations}

\label{sec:limitations}

Our study demonstrates the promise of automated mechanism discovery, but limitations remain.
First, the full evolutionary loop is still computationally intensive, since each candidate requires end-to-end training and standardized evaluation; under a fixed budget, this constrains both the breadth of explored candidates and the number of controlled follow-up trials.
Second, empirical evidence is concentrated on mathematical reasoning benchmarks and a limited range of model settings, so the transferability of the discovered mechanisms to broader regimes (\eg, open-domain dialogue, code generation, and tool-use tasks) remains to be established.
Third, although lineage analysis, training dynamics, and reflective diagnoses support mechanism-level interpretations, these interpretations should still be treated as evidence-grounded hypotheses rather than controlled causal identification of which component is responsible for each gain.
Future work should focus on improving search efficiency, broadening cross-task and cross-scale validation, and developing stronger causal evaluation protocols.

\section*{Ethics Statement}

We confirm that all authors are aware of the ACL Code of Ethics and commit to conducting this research in accordance with responsible research practices.

\paragraph{Data Use and Privacy.}
Our experiments use publicly available mathematical reasoning benchmarks (\eg, AIME, AMC, MATH, Minerva, and OlympiadBench) and do not involve private or personally identifiable information. The training subset and evaluation data are used under the respective dataset terms and licenses. We do not collect user-level behavioral data or deploy systems that interact with human subjects.

\paragraph{Automation Risks and Responsible Use.}
This work studies automated algorithm discovery for LLM-RL. Although our setting is focused on mathematical reasoning, automated search systems can, in principle, generate unstable or misaligned optimization strategies if applied without safeguards. To reduce this risk, we use bounded verification loops, standardized evaluation, and human oversight in the research pipeline, and we report limitations regarding generalization and causal interpretation. We do not claim that the discovered mechanisms are universally safe or optimal outside the evaluated scope.

\paragraph{Use of AI Assistants.}
LLMs are used as research assistants within the proposed framework (for proposal generation, implementation support, and reflective analysis) and for manuscript polishing. Final scientific claims, experimental decisions, and interpretation of results are made and verified by the authors.

\FloatBarrier
\bibliography{custom}

\clearpage
\appendix
\section{Appendix}
\label{sec:appendix}

\subsection{Full main-experiment results and lineage}
\label{sec:appendix:full_results}

\begin{figure*}[htbp]
  \centering
  \includegraphics[width=0.98\textwidth]{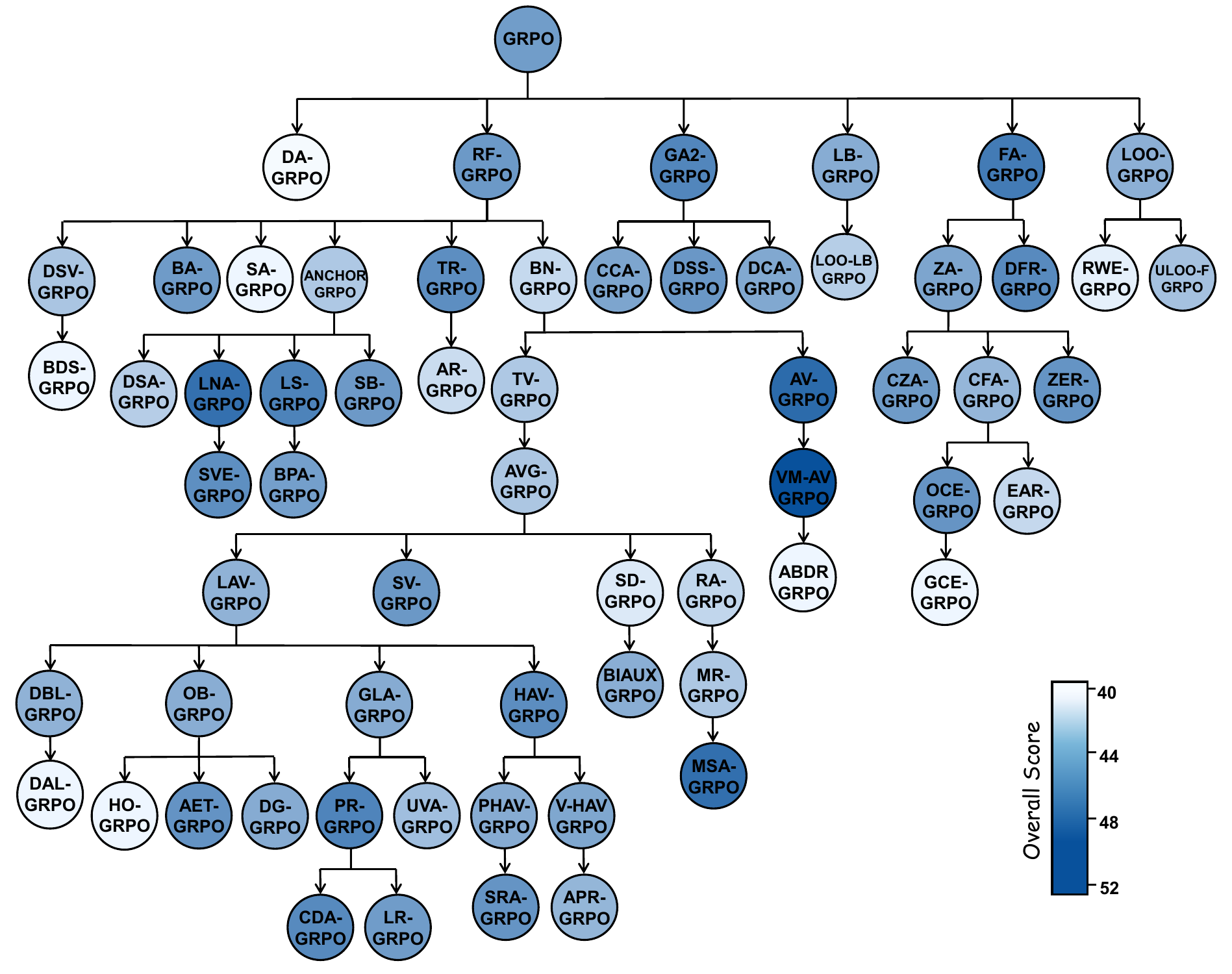}
  \caption{\textbf{Evolutionary lineage of GRPO variants in the main run.} Nodes denote algorithms, arrows denote parent-child inheritance, and node color encodes the weighted Overall (lighter is lower).}
  \label{fig:lineage}
\end{figure*}

We provide the full held-out evaluation results for all evaluated variants in \cref{tab:experiment:all_results}. To better illustrate the evolutionary paths, we also visualize the complete lineage tree in Figure~\ref{fig:lineage}.

\begin{table*}[!t]
  \centering
  \footnotesize
  \renewcommand{\arraystretch}{0.9}
  \setlength{\tabcolsep}{3pt}
  \providecommand{\resulttablewidth}{0.98\textwidth}
  \begin{tabular*}{\resulttablewidth}{@{\extracolsep{\fill}} llllllll}
    \toprule
    \textbf{Algorithm} &
    \shortstack{\textbf{AIME24}\\{\small\texttt{pass@32}}} &
    \shortstack{\textbf{AIME25}\\{\small\texttt{pass@32}}} &
    \shortstack{\textbf{AMC}\\{\small\texttt{pass@32}}} &
    \shortstack{\textbf{MATH}\\{\small\texttt{acc@1}}} &
    \shortstack{\textbf{Minerva}\\{\small\texttt{acc@1}}} &
    \shortstack{\textbf{Olympiad}\\{\small\texttt{acc@1}}} &
    \textbf{Overall} \\
    \midrule
    GRPO & 50.0 & 26.7 & 84.3 & 72.4 & 24.6 & 35.4 & 47.8 \\

VM-AV-GRPO & \score{53.3}{\textcolor{resultgreen}{+3.3}} & \score{43.3}{\textcolor{resultgreen}{+16.7}} & \score{88.0}{\textcolor{resultgreen}{+3.6}} & \score{73.6}{\textcolor{resultgreen}{+1.2}} & \score{24.3}{\textcolor{resultred}{-0.4}} & \score{35.3}{\textcolor{resultred}{-0.1}} & \score{52.5}{\textcolor{resultgreen}{+4.6}} \\
AV-GRPO & \score{56.7}{\textcolor{resultgreen}{+6.7}} & \score{36.7}{\textcolor{resultgreen}{+10.0}} & \score{81.9}{\textcolor{resultred}{-2.4}} & \score{71.0}{\textcolor{resultred}{-1.4}} & \score{26.8}{\textcolor{resultgreen}{+2.2}} & \score{35.3}{\textcolor{resultred}{-0.1}} & \score{50.9}{\textcolor{resultgreen}{+3.1}} \\
MSA-GRPO & \score{50.0}{\textcolor{resultgray}{+0.0}} & \score{43.3}{\textcolor{resultgreen}{+16.7}} & \score{84.3}{\textcolor{resultgray}{+0.0}} & \score{69.4}{\textcolor{resultred}{-3.0}} & \score{24.3}{\textcolor{resultred}{-0.4}} & \score{35.3}{\textcolor{resultred}{-0.1}} & \score{50.7}{\textcolor{resultgreen}{+2.8}} \\
LNA-GRPO & \score{56.7}{\textcolor{resultgreen}{+6.7}} & \score{33.3}{\textcolor{resultgreen}{+6.7}} & \score{84.3}{\textcolor{resultgray}{+0.0}} & \score{73.6}{\textcolor{resultgreen}{+1.2}} & \score{24.3}{\textcolor{resultred}{-0.4}} & \score{35.0}{\textcolor{resultred}{-0.4}} & \score{50.6}{\textcolor{resultgreen}{+2.7}} \\
FA-GRPO & \score{53.3}{\textcolor{resultgreen}{+3.3}} & \score{36.7}{\textcolor{resultgreen}{+10.0}} & \score{83.1}{\textcolor{resultred}{-1.2}} & \score{72.2}{\textcolor{resultred}{-0.2}} & \score{22.4}{\textcolor{resultred}{-2.2}} & \score{35.0}{\textcolor{resultred}{-0.4}} & \score{49.9}{\textcolor{resultgreen}{+2.1}} \\
LS-GRPO & \score{46.7}{\textcolor{resultred}{-3.3}} & \score{36.7}{\textcolor{resultgreen}{+10.0}} & \score{86.7}{\textcolor{resultgreen}{+2.4}} & \score{71.4}{\textcolor{resultred}{-1.0}} & \score{24.6}{\textcolor{resultgray}{+0.0}} & \score{35.7}{\textcolor{resultgreen}{+0.3}} & \score{49.4}{\textcolor{resultgreen}{+1.6}} \\
PR-GRPO & \score{53.3}{\textcolor{resultgreen}{+3.3}} & \score{36.7}{\textcolor{resultgreen}{+10.0}} & \score{80.7}{\textcolor{resultred}{-3.6}} & \score{71.2}{\textcolor{resultred}{-1.2}} & \score{23.9}{\textcolor{resultred}{-0.7}} & \score{33.2}{\textcolor{resultred}{-2.2}} & \score{49.4}{\textcolor{resultgreen}{+1.5}} \\
GA2-GRPO & \score{53.3}{\textcolor{resultgreen}{+3.3}} & \score{33.3}{\textcolor{resultgreen}{+6.7}} & \score{83.1}{\textcolor{resultred}{-1.2}} & \score{70.6}{\textcolor{resultred}{-1.8}} & \score{24.3}{\textcolor{resultred}{-0.4}} & \score{34.7}{\textcolor{resultred}{-0.7}} & \score{49.2}{\textcolor{resultgreen}{+1.4}} \\
CDA-GRPO & \score{50.0}{\textcolor{resultgray}{+0.0}} & \score{30.0}{\textcolor{resultgreen}{+3.3}} & \score{85.5}{\textcolor{resultgreen}{+1.2}} & \score{72.6}{\textcolor{resultgreen}{+0.2}} & \score{25.0}{\textcolor{resultgreen}{+0.4}} & \score{36.9}{\textcolor{resultgreen}{+1.5}} & \score{49.0}{\textcolor{resultgreen}{+1.2}} \\
DFR-GRPO & \score{46.7}{\textcolor{resultred}{-3.3}} & \score{33.3}{\textcolor{resultgreen}{+6.7}} & \score{85.5}{\textcolor{resultgreen}{+1.2}} & \score{73.4}{\textcolor{resultgreen}{+1.0}} & \score{25.7}{\textcolor{resultgreen}{+1.1}} & \score{35.1}{\textcolor{resultred}{-0.3}} & \score{49.0}{\textcolor{resultgreen}{+1.1}} \\
HAV-GRPO & \score{43.3}{\textcolor{resultred}{-6.7}} & \score{36.7}{\textcolor{resultgreen}{+10.0}} & \score{88.0}{\textcolor{resultgreen}{+3.6}} & \score{72.6}{\textcolor{resultgreen}{+0.2}} & \score{22.8}{\textcolor{resultred}{-1.8}} & \score{35.3}{\textcolor{resultred}{-0.1}} & \score{48.8}{\textcolor{resultgreen}{+0.9}} \\
TR-GRPO & \score{50.0}{\textcolor{resultgray}{+0.0}} & \score{33.3}{\textcolor{resultgreen}{+6.7}} & \score{83.1}{\textcolor{resultred}{-1.2}} & \score{72.4}{\textcolor{resultgray}{+0.0}} & \score{25.4}{\textcolor{resultgreen}{+0.7}} & \score{33.0}{\textcolor{resultred}{-2.4}} & \score{48.8}{\textcolor{resultgreen}{+0.9}} \\
SVE-LNA-GRPO & \score{50.0}{\textcolor{resultgray}{+0.0}} & \score{30.0}{\textcolor{resultgreen}{+3.3}} & \score{89.2}{\textcolor{resultgreen}{+4.8}} & \score{73.0}{\textcolor{resultgreen}{+0.6}} & \score{23.2}{\textcolor{resultred}{-1.5}} & \score{32.7}{\textcolor{resultred}{-2.7}} & \score{48.7}{\textcolor{resultgreen}{+0.9}} \\
AET-GRPO & \score{46.7}{\textcolor{resultred}{-3.3}} & \score{36.7}{\textcolor{resultgreen}{+10.0}} & \score{84.3}{\textcolor{resultgray}{+0.0}} & \score{71.2}{\textcolor{resultred}{-1.2}} & \score{23.5}{\textcolor{resultred}{-1.1}} & \score{32.7}{\textcolor{resultred}{-2.7}} & \score{48.4}{\textcolor{resultgreen}{+0.6}} \\
SRA-GRPO & \score{46.7}{\textcolor{resultred}{-3.3}} & \score{33.3}{\textcolor{resultgreen}{+6.7}} & \score{86.7}{\textcolor{resultgreen}{+2.4}} & \score{72.0}{\textcolor{resultred}{-0.4}} & \score{23.5}{\textcolor{resultred}{-1.1}} & \score{33.8}{\textcolor{resultred}{-1.6}} & \score{48.4}{\textcolor{resultgreen}{+0.6}} \\
ZER-GRPO & \score{50.0}{\textcolor{resultgray}{+0.0}} & \score{33.3}{\textcolor{resultgreen}{+6.7}} & \score{80.7}{\textcolor{resultred}{-3.6}} & \score{72.2}{\textcolor{resultred}{-0.2}} & \score{24.3}{\textcolor{resultred}{-0.4}} & \score{34.4}{\textcolor{resultred}{-1.0}} & \score{48.4}{\textcolor{resultgreen}{+0.6}} \\
OCE-GRPO & \score{46.7}{\textcolor{resultred}{-3.3}} & \score{33.3}{\textcolor{resultgreen}{+6.7}} & \score{84.3}{\textcolor{resultgray}{+0.0}} & \score{71.0}{\textcolor{resultred}{-1.4}} & \score{24.3}{\textcolor{resultred}{-0.4}} & \score{36.1}{\textcolor{resultgreen}{+0.7}} & \score{48.4}{\textcolor{resultgreen}{+0.5}} \\
DSS-GRPO & \score{53.3}{\textcolor{resultgreen}{+3.3}} & \score{26.7}{\textcolor{resultgray}{+0.0}} & \score{83.1}{\textcolor{resultred}{-1.2}} & \score{72.4}{\textcolor{resultgray}{+0.0}} & \score{23.5}{\textcolor{resultred}{-1.1}} & \score{35.4}{\textcolor{resultgray}{+0.0}} & \score{48.2}{\textcolor{resultgreen}{+0.3}} \\
SV-GRPO & \score{46.7}{\textcolor{resultred}{-3.3}} & \score{30.0}{\textcolor{resultgreen}{+3.3}} & \score{86.7}{\textcolor{resultgreen}{+2.4}} & \score{72.6}{\textcolor{resultgreen}{+0.2}} & \score{26.1}{\textcolor{resultgreen}{+1.5}} & \score{33.3}{\textcolor{resultred}{-2.1}} & \score{48.2}{\textcolor{resultgreen}{+0.3}} \\
RF-GRPO & \score{50.0}{\textcolor{resultgray}{+0.0}} & \score{33.3}{\textcolor{resultgreen}{+6.7}} & \score{80.7}{\textcolor{resultred}{-3.6}} & \score{71.2}{\textcolor{resultred}{-1.2}} & \score{23.5}{\textcolor{resultred}{-1.1}} & \score{33.9}{\textcolor{resultred}{-1.5}} & \score{48.1}{\textcolor{resultgreen}{+0.2}} \\
SB-GRPO & \score{50.0}{\textcolor{resultgray}{+0.0}} & \score{30.0}{\textcolor{resultgreen}{+3.3}} & \score{88.0}{\textcolor{resultgreen}{+3.6}} & \score{71.6}{\textcolor{resultred}{-0.8}} & \score{21.7}{\textcolor{resultred}{-2.9}} & \score{32.4}{\textcolor{resultred}{-3.0}} & \score{48.1}{\textcolor{resultgreen}{+0.2}} \\
DA-GRPO & \score{53.3}{\textcolor{resultgreen}{+3.3}} & \score{30.0}{\textcolor{resultgreen}{+3.3}} & \score{79.5}{\textcolor{resultred}{-4.8}} & \score{71.4}{\textcolor{resultred}{-1.0}} & \score{23.9}{\textcolor{resultred}{-0.7}} & \score{33.8}{\textcolor{resultred}{-1.6}} & \score{48.0}{\textcolor{resultgreen}{+0.1}} \\
BA-GRPO & \score{46.7}{\textcolor{resultred}{-3.3}} & \score{33.3}{\textcolor{resultgreen}{+6.7}} & \score{83.1}{\textcolor{resultred}{-1.2}} & \score{71.4}{\textcolor{resultred}{-1.0}} & \score{24.3}{\textcolor{resultred}{-0.4}} & \score{34.2}{\textcolor{resultred}{-1.2}} & \score{48.0}{\textcolor{resultgreen}{+0.1}} \\
LR-GRPO & \score{46.7}{\textcolor{resultred}{-3.3}} & \score{36.7}{\textcolor{resultgreen}{+10.0}} & \score{84.3}{\textcolor{resultgray}{+0.0}} & \score{70.2}{\textcolor{resultred}{-2.2}} & \score{21.7}{\textcolor{resultred}{-2.9}} & \score{32.0}{\textcolor{resultred}{-3.4}} & \score{47.9}{\textcolor{resultgreen}{+0.1}} \\
CZA-GRPO & \score{43.3}{\textcolor{resultred}{-6.7}} & \score{36.7}{\textcolor{resultgreen}{+10.0}} & \score{83.1}{\textcolor{resultred}{-1.2}} & \score{74.6}{\textcolor{resultgreen}{+2.2}} & \score{20.2}{\textcolor{resultred}{-4.4}} & \score{34.1}{\textcolor{resultred}{-1.3}} & \score{47.8}{\textcolor{resultgray}{+0.0}} \\
BPA-GRPO & \score{50.0}{\textcolor{resultgray}{+0.0}} & \score{33.3}{\textcolor{resultgreen}{+6.7}} & \score{83.1}{\textcolor{resultred}{-1.2}} & \score{71.0}{\textcolor{resultred}{-1.4}} & \score{18.8}{\textcolor{resultred}{-5.9}} & \score{34.2}{\textcolor{resultred}{-1.2}} & \score{47.7}{\textcolor{resultred}{-0.1}} \\
ZA-GRPO & \score{50.0}{\textcolor{resultgray}{+0.0}} & \score{26.7}{\textcolor{resultgray}{+0.0}} & \score{81.9}{\textcolor{resultred}{-2.4}} & \score{73.0}{\textcolor{resultgreen}{+0.6}} & \score{24.3}{\textcolor{resultred}{-0.4}} & \score{34.2}{\textcolor{resultred}{-1.2}} & \score{47.3}{\textcolor{resultred}{-0.5}} \\
CCA-GRPO & \score{50.0}{\textcolor{resultgray}{+0.0}} & \score{26.7}{\textcolor{resultgray}{+0.0}} & \score{81.9}{\textcolor{resultred}{-2.4}} & \score{71.8}{\textcolor{resultred}{-0.6}} & \score{25.0}{\textcolor{resultgreen}{+0.4}} & \score{34.7}{\textcolor{resultred}{-0.7}} & \score{47.3}{\textcolor{resultred}{-0.5}} \\
V-HAV-GRPO & \score{43.3}{\textcolor{resultred}{-6.7}} & \score{33.3}{\textcolor{resultgreen}{+6.7}} & \score{85.5}{\textcolor{resultgreen}{+1.2}} & \score{68.2}{\textcolor{resultred}{-4.2}} & \score{23.5}{\textcolor{resultred}{-1.1}} & \score{35.4}{\textcolor{resultgray}{+0.0}} & \score{47.2}{\textcolor{resultred}{-0.6}} \\
DCA-GRPO & \score{43.3}{\textcolor{resultred}{-6.7}} & \score{36.7}{\textcolor{resultgreen}{+10.0}} & \score{81.9}{\textcolor{resultred}{-2.4}} & \score{69.4}{\textcolor{resultred}{-3.0}} & \score{23.5}{\textcolor{resultred}{-1.1}} & \score{33.0}{\textcolor{resultred}{-2.4}} & \score{47.2}{\textcolor{resultred}{-0.7}} \\
DG-GRPO & \score{43.3}{\textcolor{resultred}{-6.7}} & \score{30.0}{\textcolor{resultgreen}{+3.3}} & \score{86.7}{\textcolor{resultgreen}{+2.4}} & \score{71.8}{\textcolor{resultred}{-0.6}} & \score{23.2}{\textcolor{resultred}{-1.5}} & \score{33.6}{\textcolor{resultred}{-1.8}} & \score{47.0}{\textcolor{resultred}{-0.9}} \\
PHAV-GRPO & \score{46.7}{\textcolor{resultred}{-3.3}} & \score{26.7}{\textcolor{resultgray}{+0.0}} & \score{84.3}{\textcolor{resultgray}{+0.0}} & \score{72.4}{\textcolor{resultgray}{+0.0}} & \score{24.6}{\textcolor{resultgray}{+0.0}} & \score{33.9}{\textcolor{resultred}{-1.5}} & \score{47.0}{\textcolor{resultred}{-0.9}} \\
GLA-GRPO & \score{40.0}{\textcolor{resultred}{-10.0}} & \score{33.3}{\textcolor{resultgreen}{+6.7}} & \score{86.7}{\textcolor{resultgreen}{+2.4}} & \score{71.8}{\textcolor{resultred}{-0.6}} & \score{22.1}{\textcolor{resultred}{-2.6}} & \score{34.7}{\textcolor{resultred}{-0.7}} & \score{47.0}{\textcolor{resultred}{-0.9}} \\
LB-GRPO & \score{53.3}{\textcolor{resultgreen}{+3.3}} & \score{23.3}{\textcolor{resultred}{-3.3}} & \score{83.1}{\textcolor{resultred}{-1.2}} & \score{69.0}{\textcolor{resultred}{-3.4}} & \score{23.2}{\textcolor{resultred}{-1.5}} & \score{35.1}{\textcolor{resultred}{-0.3}} & \score{46.9}{\textcolor{resultred}{-1.0}} \\
OB-GRPO & \score{53.3}{\textcolor{resultgreen}{+3.3}} & \score{26.7}{\textcolor{resultgray}{+0.0}} & \score{78.3}{\textcolor{resultred}{-6.0}} & \score{69.4}{\textcolor{resultred}{-3.0}} & \score{22.8}{\textcolor{resultred}{-1.8}} & \score{35.0}{\textcolor{resultred}{-0.4}} & \score{46.8}{\textcolor{resultred}{-1.0}} \\
BIAUX-GRPO & \score{50.0}{\textcolor{resultgray}{+0.0}} & \score{30.0}{\textcolor{resultgreen}{+3.3}} & \score{83.1}{\textcolor{resultred}{-1.2}} & \score{69.2}{\textcolor{resultred}{-3.2}} & \score{20.6}{\textcolor{resultred}{-4.0}} & \score{32.3}{\textcolor{resultred}{-3.1}} & \score{46.8}{\textcolor{resultred}{-1.1}} \\
LOO-GRPO & \score{53.3}{\textcolor{resultgreen}{+3.3}} & \score{16.7}{\textcolor{resultred}{-10.0}} & \score{85.5}{\textcolor{resultgreen}{+1.2}} & \score{72.8}{\textcolor{resultgreen}{+0.4}} & \score{24.3}{\textcolor{resultred}{-0.4}} & \score{35.4}{\textcolor{resultgray}{+0.0}} & \score{46.7}{\textcolor{resultred}{-1.1}} \\
LAV-GRPO & \score{50.0}{\textcolor{resultgray}{+0.0}} & \score{33.3}{\textcolor{resultgreen}{+6.7}} & \score{79.5}{\textcolor{resultred}{-4.8}} & \score{67.6}{\textcolor{resultred}{-4.8}} & \score{21.0}{\textcolor{resultred}{-3.7}} & \score{31.0}{\textcolor{resultred}{-4.4}} & \score{46.5}{\textcolor{resultred}{-1.3}} \\
DBL-GRPO & \score{43.3}{\textcolor{resultred}{-6.7}} & \score{33.3}{\textcolor{resultgreen}{+6.7}} & \score{80.7}{\textcolor{resultred}{-3.6}} & \score{71.0}{\textcolor{resultred}{-1.4}} & \score{22.8}{\textcolor{resultred}{-1.8}} & \score{33.3}{\textcolor{resultred}{-2.1}} & \score{46.5}{\textcolor{resultred}{-1.3}} \\
APR-GRPO & \score{43.3}{\textcolor{resultred}{-6.7}} & \score{26.7}{\textcolor{resultgray}{+0.0}} & \score{83.1}{\textcolor{resultred}{-1.2}} & \score{71.2}{\textcolor{resultred}{-1.2}} & \score{26.1}{\textcolor{resultgreen}{+1.5}} & \score{35.0}{\textcolor{resultred}{-0.4}} & \score{46.3}{\textcolor{resultred}{-1.5}} \\
CFA-GRPO & \score{33.3}{\textcolor{resultred}{-16.7}} & \score{33.3}{\textcolor{resultgreen}{+6.7}} & \score{85.5}{\textcolor{resultgreen}{+1.2}} & \score{75.0}{\textcolor{resultgreen}{+2.6}} & \score{24.3}{\textcolor{resultred}{-0.4}} & \score{34.8}{\textcolor{resultred}{-0.6}} & \score{46.3}{\textcolor{resultred}{-1.6}} \\
UVA-GRPO & \score{46.7}{\textcolor{resultred}{-3.3}} & \score{20.0}{\textcolor{resultred}{-6.7}} & \score{86.7}{\textcolor{resultgreen}{+2.4}} & \score{73.0}{\textcolor{resultgreen}{+0.6}} & \score{23.2}{\textcolor{resultred}{-1.5}} & \score{33.5}{\textcolor{resultred}{-1.9}} & \score{45.8}{\textcolor{resultred}{-2.1}} \\
ULOO-F-GRPO & \score{36.7}{\textcolor{resultred}{-13.3}} & \score{40.0}{\textcolor{resultgreen}{+13.3}} & \score{77.1}{\textcolor{resultred}{-7.2}} & \score{70.4}{\textcolor{resultred}{-2.0}} & \score{20.2}{\textcolor{resultred}{-4.4}} & \score{34.1}{\textcolor{resultred}{-1.3}} & \score{45.6}{\textcolor{resultred}{-2.2}} \\
DSV-GRPO & \score{43.3}{\textcolor{resultred}{-6.7}} & \score{30.0}{\textcolor{resultgreen}{+3.3}} & \score{80.7}{\textcolor{resultred}{-3.6}} & \score{69.2}{\textcolor{resultred}{-3.2}} & \score{23.2}{\textcolor{resultred}{-1.5}} & \score{31.6}{\textcolor{resultred}{-3.9}} & \score{45.4}{\textcolor{resultred}{-2.5}} \\
AVG-GRPO & \score{43.3}{\textcolor{resultred}{-6.7}} & \score{30.0}{\textcolor{resultgreen}{+3.3}} & \score{83.1}{\textcolor{resultred}{-1.2}} & \score{68.8}{\textcolor{resultred}{-3.6}} & \score{21.0}{\textcolor{resultred}{-3.7}} & \score{31.3}{\textcolor{resultred}{-4.1}} & \score{45.3}{\textcolor{resultred}{-2.6}} \\
MR-GRPO & \score{46.7}{\textcolor{resultred}{-3.3}} & \score{30.0}{\textcolor{resultgreen}{+3.3}} & \score{79.5}{\textcolor{resultred}{-4.8}} & \score{67.8}{\textcolor{resultred}{-4.6}} & \score{21.3}{\textcolor{resultred}{-3.3}} & \score{30.7}{\textcolor{resultred}{-4.7}} & \score{45.2}{\textcolor{resultred}{-2.6}} \\
TV-GRPO & \score{40.0}{\textcolor{resultred}{-10.0}} & \score{33.3}{\textcolor{resultgreen}{+6.7}} & \score{78.3}{\textcolor{resultred}{-6.0}} & \score{71.0}{\textcolor{resultred}{-1.4}} & \score{21.3}{\textcolor{resultred}{-3.3}} & \score{32.7}{\textcolor{resultred}{-2.7}} & \score{45.2}{\textcolor{resultred}{-2.7}} \\
ANCHOR-GRPO & \score{43.3}{\textcolor{resultred}{-6.7}} & \score{26.7}{\textcolor{resultgray}{+0.0}} & \score{79.5}{\textcolor{resultred}{-4.8}} & \score{72.4}{\textcolor{resultgray}{+0.0}} & \score{23.2}{\textcolor{resultred}{-1.5}} & \score{32.7}{\textcolor{resultred}{-2.7}} & \score{45.2}{\textcolor{resultred}{-2.7}} \\
DSA-GRPO & \score{46.7}{\textcolor{resultred}{-3.3}} & \score{23.3}{\textcolor{resultred}{-3.3}} & \score{75.9}{\textcolor{resultred}{-8.4}} & \score{73.6}{\textcolor{resultgreen}{+1.2}} & \score{23.2}{\textcolor{resultred}{-1.5}} & \score{33.0}{\textcolor{resultred}{-2.4}} & \score{44.9}{\textcolor{resultred}{-3.0}} \\
LOO-LB-GRPO & \score{43.3}{\textcolor{resultred}{-6.7}} & \score{26.7}{\textcolor{resultgray}{+0.0}} & \score{79.5}{\textcolor{resultred}{-4.8}} & \score{70.4}{\textcolor{resultred}{-2.0}} & \score{22.8}{\textcolor{resultred}{-1.8}} & \score{32.9}{\textcolor{resultred}{-2.5}} & \score{44.8}{\textcolor{resultred}{-3.0}} \\
RA-GRPO & \score{40.0}{\textcolor{resultred}{-10.0}} & \score{26.7}{\textcolor{resultgray}{+0.0}} & \score{84.3}{\textcolor{resultgray}{+0.0}} & \score{69.8}{\textcolor{resultred}{-2.6}} & \score{21.0}{\textcolor{resultred}{-3.7}} & \score{31.7}{\textcolor{resultred}{-3.7}} & \score{44.4}{\textcolor{resultred}{-3.5}} \\
EAR-GRPO & \score{36.7}{\textcolor{resultred}{-13.3}} & \score{26.7}{\textcolor{resultgray}{+0.0}} & \score{80.7}{\textcolor{resultred}{-3.6}} & \score{72.2}{\textcolor{resultred}{-0.2}} & \score{22.8}{\textcolor{resultred}{-1.8}} & \score{34.4}{\textcolor{resultred}{-1.0}} & \score{44.2}{\textcolor{resultred}{-3.7}} \\
BN-GRPO & \score{40.0}{\textcolor{resultred}{-10.0}} & \score{26.7}{\textcolor{resultgray}{+0.0}} & \score{77.1}{\textcolor{resultred}{-7.2}} & \score{72.4}{\textcolor{resultgray}{+0.0}} & \score{22.1}{\textcolor{resultred}{-2.6}} & \score{33.9}{\textcolor{resultred}{-1.5}} & \score{44.2}{\textcolor{resultred}{-3.7}} \\
AR-GRPO & \score{40.0}{\textcolor{resultred}{-10.0}} & \score{23.3}{\textcolor{resultred}{-3.3}} & \score{80.7}{\textcolor{resultred}{-3.6}} & \score{71.6}{\textcolor{resultred}{-0.8}} & \score{23.2}{\textcolor{resultred}{-1.5}} & \score{33.2}{\textcolor{resultred}{-2.2}} & \score{44.0}{\textcolor{resultred}{-3.9}} \\
SD-GRPO & \score{33.3}{\textcolor{resultred}{-16.7}} & \score{26.7}{\textcolor{resultgray}{+0.0}} & \score{81.9}{\textcolor{resultred}{-2.4}} & \score{71.2}{\textcolor{resultred}{-1.2}} & \score{22.8}{\textcolor{resultred}{-1.8}} & \score{31.9}{\textcolor{resultred}{-3.6}} & \score{43.2}{\textcolor{resultred}{-4.7}} \\
RWE-GRPO & \score{36.7}{\textcolor{resultred}{-13.3}} & \score{20.0}{\textcolor{resultred}{-6.7}} & \score{77.1}{\textcolor{resultred}{-7.2}} & \score{72.8}{\textcolor{resultgreen}{+0.4}} & \score{24.3}{\textcolor{resultred}{-0.4}} & \score{35.3}{\textcolor{resultred}{-0.1}} & \score{42.7}{\textcolor{resultred}{-5.1}} \\
BDS-GRPO & \score{36.7}{\textcolor{resultred}{-13.3}} & \score{23.3}{\textcolor{resultred}{-3.3}} & \score{75.9}{\textcolor{resultred}{-8.4}} & \score{71.4}{\textcolor{resultred}{-1.0}} & \score{21.3}{\textcolor{resultred}{-3.3}} & \score{32.7}{\textcolor{resultred}{-2.7}} & \score{42.2}{\textcolor{resultred}{-5.6}} \\
ABDR-GRPO & \score{33.3}{\textcolor{resultred}{-16.7}} & \score{20.0}{\textcolor{resultred}{-6.7}} & \score{78.3}{\textcolor{resultred}{-6.0}} & \score{72.4}{\textcolor{resultgray}{+0.0}} & \score{22.4}{\textcolor{resultred}{-2.2}} & \score{34.1}{\textcolor{resultred}{-1.3}} & \score{41.7}{\textcolor{resultred}{-6.1}} \\
DAL-GRPO & \score{16.7}{\textcolor{resultred}{-33.3}} & \score{26.7}{\textcolor{resultgray}{+0.0}} & \score{55.4}{\textcolor{resultred}{-28.9}} & \score{70.8}{\textcolor{resultred}{-1.6}} & \score{22.1}{\textcolor{resultred}{-2.6}} & \score{35.1}{\textcolor{resultred}{-0.3}} & \score{36.2}{\textcolor{resultred}{-11.7}} \\
GAA-GRPO & \score{20.0}{\textcolor{resultred}{-30.0}} & \score{10.0}{\textcolor{resultred}{-16.7}} & \score{63.9}{\textcolor{resultred}{-20.5}} & \score{71.2}{\textcolor{resultred}{-1.2}} & \score{27.6}{\textcolor{resultgreen}{+2.9}} & \score{35.0}{\textcolor{resultred}{-0.4}} & \score{35.6}{\textcolor{resultred}{-12.2}} \\
SA-GRPO & \score{20.0}{\textcolor{resultred}{-30.0}} & \score{13.3}{\textcolor{resultred}{-13.3}} & \score{54.2}{\textcolor{resultred}{-30.1}} & \score{73.2}{\textcolor{resultgreen}{+0.8}} & \score{23.9}{\textcolor{resultred}{-0.7}} & \score{33.9}{\textcolor{resultred}{-1.5}} & \score{34.5}{\textcolor{resultred}{-13.4}} \\
HO-GRPO & \score{26.7}{\textcolor{resultred}{-23.3}} & \score{6.7}{\textcolor{resultred}{-20.0}} & \score{63.9}{\textcolor{resultred}{-20.5}} & \score{67.2}{\textcolor{resultred}{-5.2}} & \score{17.3}{\textcolor{resultred}{-7.4}} & \score{33.2}{\textcolor{resultred}{-2.2}} & \score{33.9}{\textcolor{resultred}{-14.0}} \\
GCE-GRPO & \score{0.0}{\textcolor{resultred}{-50.0}} & \score{0.0}{\textcolor{resultred}{-26.7}} & \score{1.2}{\textcolor{resultred}{-83.1}} & \score{53.0}{\textcolor{resultred}{-19.4}} & \score{15.1}{\textcolor{resultred}{-9.6}} & \score{24.4}{\textcolor{resultred}{-11.0}} & \score{14.1}{\textcolor{resultred}{-33.8}}

    \\
    \bottomrule
  \end{tabular*}
  \vspace{0.5ex}
  \caption{Full held-out reasoning evaluation results for all evaluated variants (percentage points). GRPO (first row) is the baseline. Small colored numbers denote \(\Delta\) vs.\ GRPO (green: improvement; red: regression). ``Overall'' is the fixed weighted average used by the evolutionary selector (weights in \cref{sec:exp:setup}). ``Olympiad'' denotes OlympiadBench.}
  \label{tab:experiment:all_results}
\end{table*}

\subsection{Detailed Algorithm Descriptions (Representative Subset)}
\label{app:algorithms}

Due to space constraints, we do not enumerate the full evolutionary catalog here.
Instead, we summarize representative improvement and failure cases from the latest experiments.
For each algorithm, we report \textbf{motivation}, \textbf{method (with key equations)}, and \textbf{results}.
AIME24, AIME25, and AMC use \(\mathrm{pass}@32\); MATH/Minerva/OlympiadBench use \(\mathrm{acc}@1\); Overall is consistent with the main text.

\subsubsection{GRPO Baseline}
\subsubsection*{GRPO}
\textbf{Motivation.} Under sparse correctness rewards, we need a stable relative learning signal without introducing a value network.\\
\textbf{Method.} For each prompt, GRPO samples \(G\) responses and computes group-relative advantages:
\begin{align*}
\mu_g &= \frac{1}{G}\sum_{j=1}^{G} r_j, \\
\sigma_g &= \sqrt{\frac{1}{G}\sum_{j=1}^{G}(r_j-\mu_g)^2+\epsilon}, \\
A_i &= \frac{r_i-\mu_g}{\sigma_g}.
\end{align*}
The policy is then updated with a PPO-style objective plus KL regularization (divergence to a fixed reference policy).\\
\textbf{Results.} Overall \(47.8\), AIME24 \(50.0\), AIME25 \(26.7\), AMC \(84.3\), MATH \(72.4\), Minerva \(24.6\), OlympiadBench \(35.4\).

\subsubsection{Main Experiment: Representative Improving Algorithms}

This representative subset focuses on the main improving and regressive cases needed for interpretation, rather than exhaustively mirroring the compact main-table mechanism-family grouping.
Specifically, AV-GRPO and VM-AV-GRPO fall under \textbf{Core Signal Modeling}, FA-GRPO and DFR-GRPO under \textbf{Failure-side Control and Failure Reshaping}, and BN-GRPO serves as a representative \textbf{Stabilization and Constraint Control} case.
MSA-GRPO is included here as an additional high-performing variant rather than as one of the main-table exemplars for the mechanism-family grouping, while the \textbf{Signal Routing and Conditional Normalization} family is analyzed in more detail in \cref{app:detailed_analysis}.

\subsubsection*{AV-GRPO}
\textbf{Motivation.} Proposal analysis suggests that batch-level scaling in BN-GRPO improves stability but can dilute sparse hard-task success signals.\\
\textbf{Method.} AV-GRPO replaces empirical scaling with analytic-variance scaling and stricter format constraints:
\begin{align*}
\mu_g &= \frac{1}{G}\sum_i R_i, \\
\mu_{2,g} &= \frac{1}{G}\sum_i R_i^2, \\
\sigma_{\text{analytic}} &= \sqrt{\mu_{2,g}-\mu_g^2},
\end{align*}
\[
A_i=\frac{R_i-\mu_g}{\max(\sigma_{\text{analytic}},\sigma_{\min})}.
\]
For binary rewards, this reduces to \(\sigma=\sqrt{\mu(1-\mu)}\), providing a more adaptive scale in rare-success regimes.
In the proposal, this scaling-centered redesign is paired with a strict format penalty and a lower entropy coefficient, so the empirical gain should be interpreted as the effect of a small set of coordinated changes rather than a single isolated substitution.\\
\textbf{Results.} Overall \(50.9\), AIME24 \(56.7\), AIME25 \(36.7\), with gains over GRPO.

\subsubsection*{VM-AV-GRPO}
\textbf{Motivation.} The training proposal identifies a residual mismatch in AV-GRPO: mixing invalid and valid samples in shared statistics can produce false-positive incentives for valid-but-wrong outputs.\\
\textbf{Method.} VM-AV-GRPO applies validity masking with Bayesian-smoothed valid-subset statistics (computed on the format-valid subset only), together with an explicit floor penalty for invalid outputs:
\begin{align*}
\mu_{\text{valid}} &= \frac{\sum_{i\in V}R_i+\alpha}{|V|+\alpha+\beta}, \\
\sigma_{\text{valid}} &= \sqrt{\mu_{\text{valid}}(1-\mu_{\text{valid}})},
\end{align*}
\[
A_i=
\begin{cases}
\dfrac{R_i-\mu_{\text{valid}}}{\max(\sigma_{\text{valid}},\sigma_{\text{floor}})}, & i\in V\\
A_{\text{floor}}, & i\notin V.
\end{cases}
\]
It further applies validity-gated length normalization (only on valid samples) to negative valid advantages:
\begin{align*}
L_{\text{rel}} &= \mathrm{clip}\!\left(\frac{L_i}{\bar L_{\text{valid}}},0.5,2.0\right), \\
A_i^{\text{final}} &=
\begin{cases}
A_i/L_{\text{rel}}, & i\in V\land A_i<0\\
A_i, & \text{otherwise}.
\end{cases}
\end{align*}
\textbf{Results.} Overall \(52.5\), AIME25 \(43.3\), AMC \(88.0\); higher than AV-GRPO on AMC (\(81.9\)).

\subsubsection*{MSA-GRPO}
\textbf{Motivation.} The proposal separates high-information mixed groups from low-information uniform groups, arguing they should not drive updates at equal magnitude.\\
\textbf{Method.} MSA-GRPO introduces regime-aware anchor selection and scaling (Mixed: both correct and incorrect samples; Uniform: all same outcome):
\[
\sigma_{\text{anchor}}=
\begin{cases}
\sigma_{\text{outcome}}, & \sigma_{\text{outcome}}>\epsilon\\
\sigma_{\text{total}}, & \text{otherwise}
\end{cases}
\]
\[
A_{\text{scaled}}=A_{\text{raw}}\cdot
\begin{cases}
1.0, & \text{Mixed}\\
\alpha_{\text{uniform}}, & \text{Uniform}
\end{cases}
\]
with multiplicative length scaling:
\begin{align*}
\text{ratio}_i &= \frac{L_i}{\bar L_{\text{group}}}, \\
A_{\text{final}} &=
\begin{cases}
A_{\text{scaled}}\cdot \text{ratio}_i^{\lambda}, & A_{\text{scaled}}>0\\
A_{\text{scaled}}\cdot \text{ratio}_i^{-\lambda}, & A_{\text{scaled}}<0.
\end{cases}
\end{align*}
\textbf{Results.} Overall \(50.7\), AIME25 \(43.3\).

\subsubsection*{FA-GRPO}
\textbf{Motivation.} Proposal evidence highlights all-fail stagnation under standard GRPO, motivating explicit signal shaping in all-fail groups.\\
\textbf{Method.} FA-GRPO uses a two-regime advantage:
\[
A_{\text{final}}=
\begin{cases}
\mathrm{clip}\!\left(\dfrac{R_i-\mu_g}{\sigma_g+\epsilon},-3,3\right), & R_{\max}>0\\
-2.0, & \text{invalid format}\\
-0.5, & \text{valid but wrong}.
\end{cases}
\]
This design separates structural errors from reasoning failures with different penalty magnitudes.\\
\textbf{Results.} Overall \(49.9\), AIME25 \(36.7\).

\subsubsection*{DFR-GRPO}
\textbf{Motivation.} Post-run diagnosis of FA-GRPO suggests that fixed penalties can destabilize formatting when mixed directly with normalized reasoning advantages, and that uniform negative penalties in all-fail groups create directionless repulsion rather than directional correction.\\
\textbf{Method.} DFR-GRPO decouples format control and reasoning quality into independent advantage streams:
\[
A_i^{\text{total}} = A_i^{\text{reason}} + \lambda_{\text{fmt}} A_i^{\text{format}},
\]
where \(A_i^{\text{format}}\) tracks format compliance separately from reasoning quality.
In all-fail groups, the reasoning stream no longer uses a uniform constant penalty; instead, it assigns a zero-centered signal derived from token entropy, so that more informative failures are compared against less informative ones rather than being pushed equally.
This turns failure-side updates from undirected repulsion into directional comparison among failures.\\
\textbf{Results.} Overall \(49.0\), AIME25 \(33.3\), AMC \(85.5\).
Relative to FA-GRPO, this revision primarily regularizes the failure-side signal path, while incurring trade-offs on some hard benchmarks.

\subsubsection{Main Experiment: Stabilization and Regression Cases}

\subsubsection*{BN-GRPO}
\textbf{Motivation.} BN-GRPO aims to suppress noise amplification via group-centering and batch-scaling, while moving KL/entropy terms (entropy encourages exploration) into the loss.\\
\textbf{Method.}
\begin{align*}
A_{\text{raw},ij} &= R_{ij}-\mu_i, \\
\sigma_{\text{batch}} &= \sqrt{\frac{1}{B}\sum_{i,j}A_{\text{raw},ij}^2}, \\
A_{\text{final},ij} &= \mathrm{clip}\!\left(\frac{A_{\text{raw},ij}}{\sigma_{\text{batch}}+\epsilon},-3,3\right),
\end{align*}
\[
L(\theta)=\mathbb{E}\!\left[L^{\text{CLIP}}(\theta)-\beta_{\text{KL}}D_{\text{KL}}+\beta_{\text{ent}}H(\pi)\right].
\]
\textbf{Results.} Overall \(44.2\), AIME25 \(26.7\).
In this stage, the stabilization objective does not translate into hard-task gains.

\subsubsection*{SA-GRPO}
\textbf{Motivation.} SA-GRPO uses fixed scaling and a dual-anchor baseline to avoid extreme advantages in low-variance groups.\\
\textbf{Method.}
\begin{align*}
\mu_{\text{anchor}} &= \alpha\mu_{\text{group}}+(1-\alpha)\mu_{\text{batch}}, \\
A_{ij} &= \frac{R_{ij}-\mu_{\text{anchor}}}{\sigma_{\text{fixed}}},
\end{align*}
with decoupled regularization in loss:
\[
L_{\text{total}}=L_{\text{CLIP}}-\beta_{\text{KL}}D_{\text{KL}}+\gamma_{\text{Ent}}H(\pi).
\]
\textbf{Results.} Overall \(34.5\), AIME25 \(13.3\), representing a regression case.

\subsubsection{Response-Length Compression Experiment: Success and Failure}

\subsubsection*{DACE-GRPO (Success)}
\textbf{Motivation.} Compress response length while preserving hard-task correctness, avoiding ``compress-first, regress-later'' behavior.\\
\textbf{Method.} Difficulty-aware reweighting is applied via group pass rate (the fraction of correct samples within a prompt group):
\[
\text{PassRate}_i=\frac{1}{G}\sum_{j=1}^{G}\mathbb{I}(r_{ij}>0.5),
\]
\begin{align*}
A_{\text{acc}}^{\text{new}} &= A_{\text{acc}}\bigl(1+\alpha(1-\text{PassRate}_i)\bigr), \\
A_{\text{eff}}^{\text{new}} &= A_{\text{eff}}\cdot \text{PassRate}_i,
\end{align*}
\[
A_{\text{total}}=A_{\text{acc}}^{\text{new}}+\beta A_{\text{eff}}^{\text{new}}\cdot \mathbb{I}(\text{Correct}_i).
\]
A_{\text{eff}} is computed from the z-scored negative response lengths within the correct subset, so the compression signal only acts among correct samples and is then further gated by group pass rate.
An annealed entropy target \(H_{\text{target}}(t)\) further supports exploration early and consolidation later.\\
\textbf{Results.} Overall \(51.7\) (\(+3.9\)), mean length \(473.6\to335.7\) (ratio \(0.709\)); AIME25 \(40.0\), AMC \(86.7\).

\subsubsection*{CAG-GRPO (Success)}
\textbf{Motivation.} Prevent length collapse driven by disproportionate gradients from overly short solutions, by reformulating efficiency as a cost term rather than a positive bonus.\\
\textbf{Method.}
\begin{align*}
A_{\text{eff}} &= \min\!\left(\frac{\mu_L-L_i}{\mu_L},\,0\right), \\
\beta &= \beta_{\text{base}}\cdot SR^2, \\
A_{\text{total}} &= A_{\text{acc}}+\beta A_{\text{eff}}\cdot \mathbb{I}(\text{Correct}_i).
\end{align*}
Here \(\mu_L\) is computed over the correct subset within a prompt group.
The non-positive efficiency adjustment removes explicit bonuses for being shorter than the mean, reducing the risk of over-compression.\\
\textbf{Results.} Overall \(49.5\) (\(+1.6\)), mean length ratio \(0.854\); AIME25 \(40.0\), AMC \(84.3\).

\subsubsection*{DCBE-GRPO (Failure)}
\textbf{Motivation.} DCBE-GRPO introduces a relative-length-improvement efficiency term and difficulty gating, aiming for stable compression without accuracy collapse.\\
\textbf{Method.}
\begin{align*}
RLI_i &= \frac{\mu_L-L_i}{\mu_L+\epsilon}, \\
\alpha_{\text{diff}} &= \left(\frac{N_{\text{correct}}}{N_{\text{total}}}\right)^2,
\end{align*}
\[
A_{\text{total}}=A_{\text{acc}}+\lambda\cdot \alpha_{\text{diff}}\cdot RLI_i\cdot \mathbb{I}(\text{Correct}_i).
\]
\noindent where \(\mu_L\) denotes the mean length computed over the correct subset within a prompt group.
\noindent In the implemented recipe, this efficiency path is paired with a dynamic entropy controller.
\textbf{Results.} Overall \(27.8\) (\(-20.0\)), mean length ratio \(1.333\) (length increases); AIME25 \(6.7\), AMC \(45.8\).
This therefore serves as a negative case for the combined design, where the RLI-style efficiency term alone does not explain the observed failure.

\subsubsection{Summary}
Across these representative cases, improvements are associated with finer signal hierarchy design (\eg, validity separation and difficulty-aware reweighting), rather than stronger penalties alone.
Failure cases indicate that stabilization or efficiency objectives can still degrade hard reasoning when not coordinated with task difficulty and signal structure.

\subsection{Case Study of Evolutionary Discovery Workflow}
\label{app:case_study}

To illustrate how the framework accumulates and reuses evidence, we present a detailed trace of a single evolutionary lineage, visualized in Figure~\ref{fig:lineage} and grounded in the full held-out results in \cref{tab:experiment:all_results}.
This case study shows how stronger variants emerge under a shared empirical protocol, and how failures, revisions, and follow-up hypotheses are linked across iterations.

The chosen lineage follows the path: GRPO $\to$ RF-GRPO $\to$ BN-GRPO $\to$ AV-GRPO $\to$ VM-AV-GRPO.
This path is selected because it captures two informative phases of the search process.
In the mechanism-family grouping used in \cref{sec:exp:results}, it begins with \textbf{Stabilization and Constraint Control} (RF-GRPO, BN-GRPO) and culminates in \textbf{Core Signal Modeling} (AV-GRPO, VM-AV-GRPO):
\begin{enumerate}
    \item \textbf{A Performance Jump:} Starting from insufficient use of failure-group signals and unstable sparse-success amplification, the lineage passes through stabilization attempts and a regression before reaching analytic-variance scaling, which is associated with a clear performance gain.
    \item \textbf{A Follow-up Correction:} After that gain, the system continues to diagnose a residual false-positive incentive on valid-but-wrong samples and corrects it with validity masking, validity-gated length normalization, and adaptive entropy control.
\end{enumerate}

\subsubsection*{Root: GRPO Baseline}
\textbf{Status:} Baseline (Overall: 47.8, AIME25: 26.7\%)

The lineage begins with standard GRPO.
While competitive on broader benchmark coverage, the framework identifies a clear bottleneck on sparse-success benchmarks such as AIME25.
\begin{itemize}[leftmargin=*, nosep]
    \item \textbf{Diagnosed Deficiency:} Under sparse success, the training signal faces two structural challenges: (i) when only a few samples succeed in a group, the within-group variance can be tiny, inflating advantage scale and inducing update shocks and distribution collapse; and (ii) in all-wrong groups, rewards provide little discriminative relative structure, yielding weak directional learning signals.
    This makes it difficult to balance exploration with consolidating rare successful trajectories on hard prompts.
\end{itemize}

\subsubsection*{Attempt 1: Reference Fallback and Damped Normalization (RF-GRPO)}
\textbf{Status:} Targeted gain (Overall: 48.1, AIME25: 26.7\% $\to$ 33.3\%)

To mitigate sparse-success shocks and failure-mode lock-in, the framework proposes RF-GRPO.
\begin{itemize}[leftmargin=*, nosep]
    \item \textbf{Mechanism:} It introduces damped advantage normalization to cap outlier updates, performance-gated KL fallback to the reference policy for all-fail groups, and hierarchical format rewards to provide denser structural feedback.
    \item \textbf{Outcome:} AIME25 improves over the baseline, but the overall gain remains limited and failure-group noise can still interfere with learning on the hardest problems.
    \item \textbf{Diagnosis:} In this setting, the stabilizing controls improve training stability, but a remaining issue persists: if failure-group noise is normalized without distinction, signal contamination can still hinder hard-task learning.
\end{itemize}

\subsubsection*{Stabilization Attempt: Batch-Level Scaling (BN-GRPO)}
\textbf{Status:} Stable but regressive (Overall: 44.2, AIME25: 26.7\%)

To reduce failure-noise contamination, the framework proposes BN-GRPO, decoupling group-centering and batch-level scaling, while moving KL/entropy regularization explicitly into the optimization objective.
\begin{itemize}[leftmargin=*, nosep]
    \item \textbf{Outcome:} While motivated by improved stability, evaluation shows a regression in aggregate performance and no improvement over the baseline on AIME25.
    \item \textbf{Takeaway:} Stabilization alone is not equivalent to hard-task progress.
    If the advantage scale is not explicitly aligned with success rarity, suppressing drift can also attenuate the positive signal needed for rare-success consolidation.
\end{itemize}

\subsubsection*{The Success Leap: Analytic-Variance Scaling (AV-GRPO)}
\textbf{Status:} \textbf{Breakthrough} (Overall: 50.9, AIME25: 26.7\% $\to$ \textbf{36.7\%})

Based on the diagnosis above, the framework proposes AV-GRPO, replacing empirical-scale heuristics with analytic-variance scaling for advantage normalization.
\begin{itemize}[leftmargin=*, nosep]
    \item \textbf{Mechanism 1: Analytic-Variance Advantage Scaling.} Advantages are normalized by a theoretically grounded variance term, so that rarer successes naturally produce stronger positive updates.
    \item \textbf{Mechanism 2: Stricter format constraints and conservative entropy regularization.} This reduces invalid exploration noise while preserving effective sparse-signal learning.
    \item \textbf{Evidence:} This iteration yields a clear jump on both aggregate and hard-task metrics, including an AIME24 score of \(56.7\), consistent with the view that stronger rare-signal amplification helps difficult reasoning.
\end{itemize}

\subsubsection*{The Research Loop: Validity-Masked Correction (VM-AV-GRPO)}
\textbf{Status:} Refinement gain (Overall: 52.5, AIME25: 36.7\% $\to$ 43.3\%)

After the AV-GRPO breakthrough, the framework continues diagnosing a residual mismatch: when invalid samples are included in the same statistics, \emph{valid-but-wrong} responses can be over-rewarded.
\begin{itemize}[leftmargin=*, nosep]
    \item \textbf{Mechanism:} VM-AV-GRPO applies validity masking with Bayesian-smoothed valid-subset statistics, assigns a floor penalty to invalid outputs, and combines validity-gated length normalization with adaptive entropy control.
    \item \textbf{Outcome \& Diagnosis:} The update improves Overall and AIME25 while also yielding a concurrent gain on AMC (AV-GRPO: \(81.9\) $\rightarrow$ VM-AV-GRPO: \(88.0\)). This pattern is consistent with the view that separating structural validity from reasoning quality leads to a cleaner sparse-reward signal hierarchy.
    \item \textbf{Interpretation:} This step links an empirical gain to a residual mismatch and a subsequent mechanism revision, rather than treating the first improvement as the end of the search.
\end{itemize}

\subsection{Detailed Evolutionary Trajectories and Evidence Accumulation}
\label{app:detailed_analysis}

This subsection expands the concise analysis in \cref{sec:analysis} by making the intermediate turning points, failed hypotheses, and cross-lineage transfer patterns more explicit.

\begin{figure*}[t]
  \centering
  \includegraphics[width=0.78\textwidth]{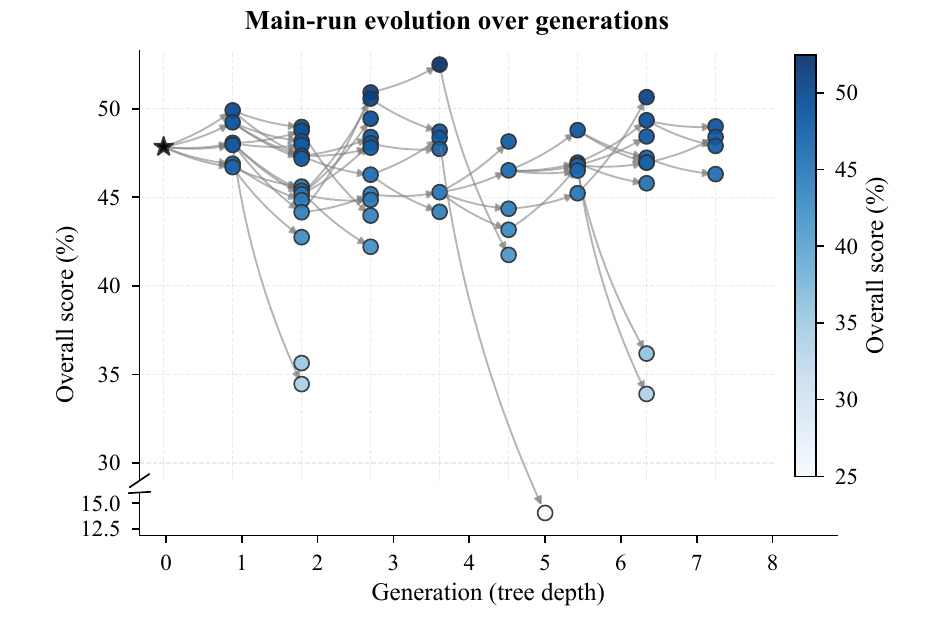}
  \caption{Complete view of the full base-branch lineage by tree depth.
  Nodes are plotted by tree depth and Overall score, and arrows indicate parent-child relations.
  Relative to the summary frontier view in \cref{fig:depth_frontier_base}, this figure makes the underlying structure explicit: deeper generations still contain exploratory failures, and several strong descendants emerge only after intermediate regressions or branch-specific refinements.}
  \label{fig:mainrun_local_optimum_base}
\end{figure*}

\subsubsection{Retrospective Evidence on Parent Retention}
\label{app:analysis:parent_retention}

Across the \(21\) comparable parent-selection rounds in the base branch, three parents are retained in each round, yielding \(63\) retained-parent decisions in total.
These retained sets cover about \(1.8\) top-level branches on average, and \(14\) rounds span at least two branches.

More importantly, a parent's score at selection and its later contribution are not equivalent.
Among the \(62\) retained-parent decisions whose descendants can be observed later in the lineage, \(43\) yield a descendant with higher Overall than the selected parent itself, with a median gain of \(+2.6\) Overall; \(27\) improve by more than \(3\) points, and \(12\) improve by more than \(5\) points.
In particular, \(27\) of the \(33\) selections whose parent is below the GRPO baseline eventually lead to descendants that surpass that baseline.

The evidence becomes sharper when we compare parents selected in the same round.
In more than half of the comparable rounds (\(11/21\)), the parent that eventually produces the strongest descendant is not the selected parent with the highest Overall score at selection time; across all comparable within-round parent pairs, \(26/50\) reverse order between Overall score at selection and eventual best-descendant score.
One representative round is as follows (numbers in parentheses denote Overall scores): ZA-GRPO (\(47.3\)), BN-GRPO (\(44.2\)), and Anchor-GRPO (\(45.2\)) are retained together; the strongest later descendant is VM-AV-GRPO (\(52.5\)) from BN-GRPO, whereas the ZA-GRPO branch peaks at \(48.4\).
Another representative round is as follows (numbers in parentheses denote Overall scores): LAV-GRPO (\(46.5\)), GA2-GRPO (\(49.2\)), and RA-GRPO (\(44.4\)) are retained together; here the selected parent with the lowest Overall score, RA-GRPO, later reaches MSA-GRPO (\(50.7\)), exceeding the GA2-GRPO branch peak of \(48.2\).
Representative cases are listed in \cref{tab:parent_retention_examples}.
These observations support the intuition behind retaining parents for future improvement rather than filtering solely by current Overall score; they constitute retrospective rather than controlled causal evidence.

\begin{table}[t]
  \centering
  \small
  \caption{Representative selected parents whose strongest descendants substantially outperform them.
  Scores are Overall.
  Several of these cases arise in rounds where a lower-scoring selected parent later outperforms the branch of an initially stronger selected parent.}
  \label{tab:parent_retention_examples}
  \begin{tabular}{lccc}
    \toprule
    Parent & Overall & Best descendant & Gain \\
    \midrule
    BN-GRPO & 44.2 & VM-AV-GRPO (52.5) & +8.3 \\
    Anchor-GRPO & 45.2 & LNA-GRPO (50.6) & +5.4 \\
    RA-GRPO & 44.4 & MSA-GRPO (50.7) & +6.3 \\
    GLA-GRPO & 47.0 & PR-GRPO (49.4) & +2.4 \\
    CFA-GRPO & 46.3 & OCE-GRPO (48.4) & +2.1 \\
    \bottomrule
  \end{tabular}
\end{table}

\subsubsection{Evolution Dynamics: Exploration and Refinement in Mechanism Space}
\label{app:analysis:trends}

The evolutionary trajectory reveals a stepwise refinement process: each generation typically introduces a relatively localized mechanism revision, or a compact bundle of coordinated changes, on top of its parent and is then filtered by the evaluation protocol.
To explain these turning points, we refer to three core training signals: (i) \textbf{advantage}, which governs the relative weight of rare successes versus failures in updates; (ii) \textbf{KL divergence}, which constrains deviation from a reference model and controls update size; and (iii) \textbf{entropy}, which helps diagnose premature collapse or excessive drift.

The traces suggest that performance jumps often coincide with restructuring and decoupling learning-signal pathways rather than simply increasing a single signal magnitude, together with recalibrating penalties on failures and stability constraints.
We trace three representative lineages that align with three of the four mechanism families in the main text; the remaining family, \textbf{Stabilization and Constraint Control}, appears as an important precursor within the first lineage:

\paragraph{(i) Core Signal Modeling (with Stabilization and Constraint Control Precursors).}
GRPO (Overall \(47.8\), AIME25 \(26.7\)) $\rightarrow$ RF-GRPO (Overall \(48.1\), AIME25 \(33.3\)) $\rightarrow$ BN-GRPO (Overall \(44.2\), AIME25 \(26.7\)) $\rightarrow$ AV-GRPO (Overall \(50.9\), AIME25 \(36.7\)) $\rightarrow$ VM-AV-GRPO (Overall \(52.5\), AIME25 \(43.3\)).
\textbf{Main transition:} This lineage refines how sparse successes are safely amplified.
The GRPO baseline suffers from entropy collapse caused by gradient shocks from rare successes, and provides no usable relative signal in all-fail groups.
RF-GRPO introduces a stabilizing control: damped normalization caps outlier advantages, dynamic reference fallback prevents deterministic failure modes, and hierarchical rewards provide dense format feedback.
BN-GRPO further diagnoses a side effect of per-group whitening: failure groups dominated by format noise can be amplified to the same magnitude as success groups, diluting the rare-success signal needed for hard tasks.
To break this bottleneck, AV-GRPO shifts to explicit rare-event modeling by anchoring normalization to analytic variance so that rarer successes can yield stronger positive updates.
VM-AV-GRPO then addresses a false-positive failure mode: when invalid samples enter the statistics, valid-but-wrong outputs are often assigned positive advantages.
It restores the strict gradient hierarchy (invalid $\ll$ valid-wrong $<$ valid-correct) via validity masking with valid-subset statistics, combining this with validity-gated length normalization and an adaptive entropy controller to further improve AIME25 and Overall, though navigating trade-offs on certain hard metrics.

\paragraph{(ii) Failure-side Control and Failure Reshaping.}
GRPO (Overall \(47.8\), AIME25 \(26.7\)) $\rightarrow$ FA-GRPO (Overall \(49.9\), AIME25 \(36.7\)) $\rightarrow$ DFR-GRPO (Overall \(49.0\), AIME25 \(33.3\)).
\textbf{Main transition:} This lineage targets all-fail stagnation and related failure modes.
FA-GRPO attempts to break the zero-gradient issue by branching on group success: it assigns hierarchical fixed penalties in all-fail groups (mild for logic errors, stronger for format errors).
However, subsequent evidence reveals two pitfalls of fixed penalties: mixing scalars with normalized advantages causes scale mismatch that destabilizes formatting, and applying uniform negative advantages acts as "directionless repulsion," often triggering drift and verbose failures.
DFR-GRPO addresses this by separating the signals: it decouples format and reasoning into two independent advantage streams, and replaces uniform repulsion with a zero-centered signal derived from token entropy.
This signal compares more informative failures against less informative ones, so the policy receives a directional gradient even when all candidates are incorrect, providing more robust learning signals in all-fail groups, albeit with trade-offs on the hardest benchmarks.

\paragraph{(iii) Signal Routing and Conditional Normalization.}
GRPO (Overall \(47.8\), AIME25 \(26.7\)) $\rightarrow$ GLA-GRPO (Overall \(47.0\), AIME25 \(33.3\)) $\rightarrow$ PR-GRPO (Overall \(49.4\), AIME25 \(36.7\)) $\rightarrow$ CDA-GRPO (Overall \(49.0\), AIME25 \(30.0\)).
\textbf{Main transition:} This lineage illustrates the pitfalls of scalarizing multiple signals too early.
While GLA-GRPO improves stability, it leaves all-fail groups with little usable learning signal.
PR-GRPO makes routing explicit by partitioning groups into solved/failed/mixed regimes: it silences mastered groups, penalizes all-fail groups, and normalizes only mixed-group residuals to improve signal fidelity.
CDA-GRPO addresses a complementary bottleneck: aggregating outcome, format, and efficiency (length) into a single scalar fundamentally causes interference, and hard-silencing solved groups creates an efficiency ``deadzone.''
To fix this, CDA-GRPO splits advantages into independent streams and conditionally normalizes efficiency only within the correct subset (``first be correct, then be short'').
It replaces brittle regime cliffs with a continuous optimistic baseline, which helps suppress short-and-wrong tendencies caused by signal conflict.
This supports a core heuristic: secondary objectives should be introduced through signal decoupling and conditional normalization so that they complement, rather than dilute, the primary correctness signal.

\subsubsection{Evidence Accumulation and Cross-Lineage Transfer}
\label{app:analysis:lineage_evidence}

Based on standardized evaluation across all \(64\) nodes, we view each iteration as a mechanism-level experiment: each proposal states an anticipated bottleneck and a corrective hypothesis, while evaluation results and key training signals (advantage/KL/entropy) provide feedback for later revisions.
Importantly, this feedback does not remain isolated within a single path.
In later iterations, negative outcomes help mark out practical design limits, while components that repeatedly correlate with gains are reused in new combinations.
We summarize three primary modes of cross-lineage knowledge transfer:

\paragraph{Negative-evidence-driven path pruning.}
Negative examples are often fed back into later proposals as algorithm-level "failure diagnoses" (\eg, the FA$\rightarrow$DFR progression in \cref{app:analysis:trends}).
For instance, FA-GRPO attempts to inject gradients into all-fail groups via fixed scalar penalties, but DFR-GRPO's diagnosis reveals that mixing fixed penalties with normalized advantages causes scale mismatch, and applying identical negative signals to all failures lacks directionality, inducing policy drift.
Consequently, later designs tended to decouple format control from reasoning quality into comparable independent streams, and to introduce zero-centered token-entropy signals in all-fail groups to eliminate directionless repulsion.

\paragraph{Horizontal diffusion of stability mechanisms.}
Stability components frequently diffuse across paths.
For example, advantage clipping and explicit KL regularization are reused across multiple representative paths (\eg, BN-GRPO, DFR-GRPO, and later variants) to suppress excessively large updates when amplifying rare successes.
Later variants incorporate these regularizers into the loss function and pair them with clipping to control drift, shifting stability control from a single global hyperparameter to more fine-grained local constraints.
This reuse reduces the overhead of exploring new algorithmic combinations, allowing more of the search budget to be spent on signal-structure changes rather than repeated recovery from unstable training.

\paragraph{Recurring empirical regularities.}
As evolution deepens, recurring heuristics across branches gradually consolidate into stable empirical rules: first, in success-scarce regimes, penalties on failures should generally avoid coupling with length to prevent a bias toward short but uninformative answers; second, secondary objectives are better introduced via "signal decoupling and conditional normalization" so that the primary signal does not overwhelm finer-grained objectives.
VM-AV-GRPO's fix for false-positive advantages on "valid-but-wrong" samples, and CDA-GRPO's "first be correct, then be short" conditional logic, both follow the same pattern: establish comparable, routable learning signals first, then introduce secondary preferences under controlled conditions.

In summary, lineage evolution can be viewed as a process of \emph{evidence accumulation}.
Negative examples mark exploration boundaries, component reuse lowers exploratory overhead, and recurring regularities give later proposals a stronger starting point.
This helps improve search efficiency and the quality of algorithm discovery under a fixed computational budget.

\subsection{Training Dynamics under Length-Compression Constraints}
\label{app:compression_dynamics}

Table~\ref{tab:compression_compact} provides a quantitative summary of all \(11\) nodes in the length-compression branch (\(1\) GRPO baseline + \(10\) evolved variants).
To keep the table readable, we report four representative dataset columns explicitly, while Overall and mean length are still computed over the full six-dataset evaluation suite.
Mean length is computed as the unweighted mean word count over the six evaluation datasets.

\begin{table*}[h]
  \centering
  \footnotesize
  \renewcommand{\arraystretch}{0.92}
  \setlength{\tabcolsep}{4pt}
  \begin{tabular*}{0.98\textwidth}{@{\extracolsep{\fill}} lrrrrrrr}
    \toprule
    \textbf{Algorithm} &
    \shortstack{\textbf{AIME24}\\{\small\texttt{pass@32}}} &
    \shortstack{\textbf{AIME25}\\{\small\texttt{pass@32}}} &
    \shortstack{\textbf{AMC}\\{\small\texttt{pass@32}}} &
    \shortstack{\textbf{MATH}\\{\small\texttt{acc@1}}} &
    \textbf{Overall} &
    \shortstack{\textbf{Mean}\\\textbf{Length}} &
    \shortstack{\textbf{Length}\\\textbf{Ratio}} \\
    \midrule
    GRPO & 50.0 & 26.7 & 84.3 & 72.4 & 47.8 & 473.6 & 1.000 \\
    DACE-GRPO & 56.7 & 40.0 & 86.7 & 72.2 & 51.7 & 335.7 & 0.709 \\
    CAG-GRPO & 46.7 & 40.0 & 84.3 & 73.0 & 49.5 & 404.4 & 0.854 \\
    DRE-GRPO & 56.7 & 36.7 & 80.7 & 69.6 & 49.4 & 318.0 & 0.671 \\
    PR-GRPO & 53.3 & 36.7 & 80.7 & 71.2 & 49.4 & 532.5 & 1.124 \\
    PAS-GRPO & 50.0 & 30.0 & 85.5 & 73.4 & 48.5 & 367.7 & 0.776 \\
    DCE-GRPO & 43.3 & 33.3 & 80.7 & 72.4 & 47.4 & 484.0 & 1.022 \\
    DQG-GRPO & 43.3 & 30.0 & 80.7 & 72.0 & 46.4 & 325.5 & 0.687 \\
    MCE-GRPO & 43.3 & 33.3 & 80.7 & 65.8 & 45.2 & 239.5 & 0.506 \\
    CAS-GRPO & 43.3 & 23.3 & 75.9 & 69.6 & 43.1 & 1092.3 & 2.306 \\
    DCBE-GRPO & 10.0 & 6.7 & 45.8 & 65.2 & 27.8 & 631.2 & 1.333 \\
    \bottomrule
  \end{tabular*}
  \caption{Quantitative summary of all \(11\) nodes in the length-compression branch.
  Mean length denotes the unweighted mean word count across the six evaluation datasets.
  Length ratio is computed relative to the GRPO baseline.}
  \label{tab:compression_compact}
\end{table*}

\begin{figure*}[h]
  \centering
  \includegraphics[width=\linewidth]{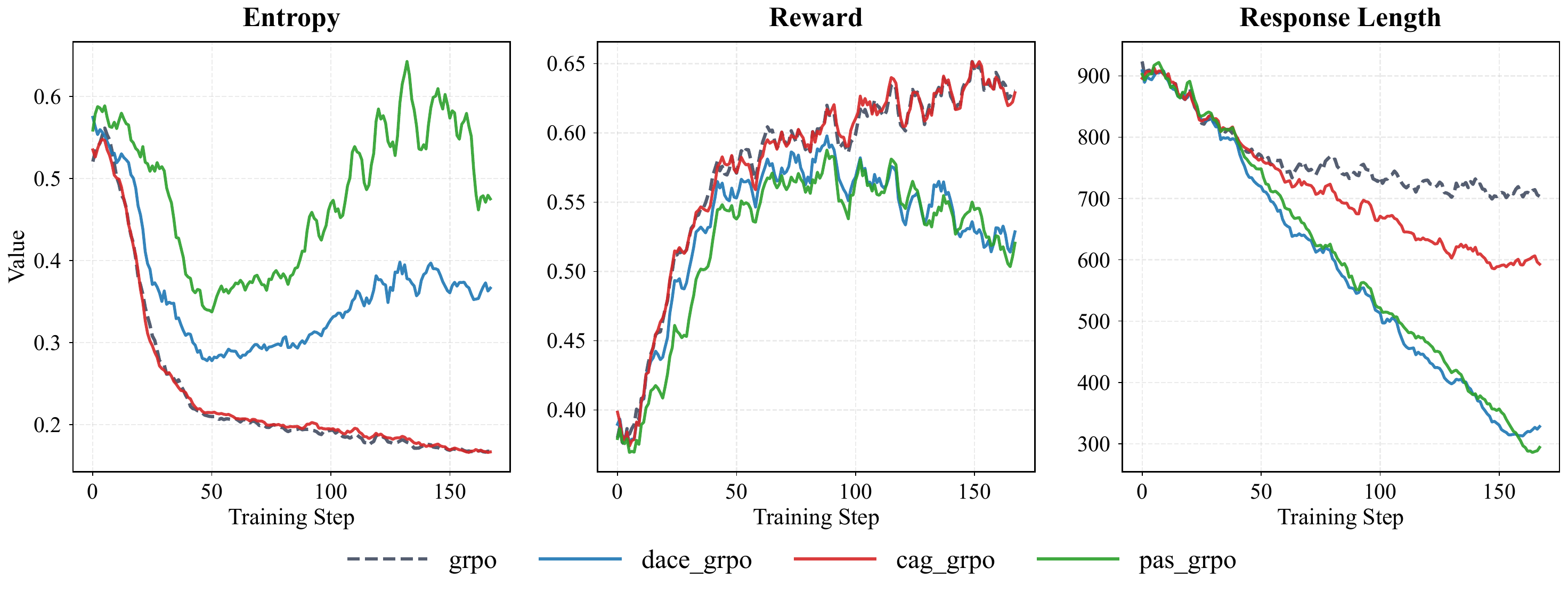}
  \caption{Training dynamics under the length-compression constraint for representative successful variants. The plots compare entropy, reward, and response length over training.}
  \label{fig:training_dynamics_compression_app}
\end{figure*}

Figure~\ref{fig:training_dynamics_compression_app} illustrates the training dynamics of representative successful variants evolved under the length-compression constraint.
Together with the failure cases summarized in the representative algorithm descriptions (\cref{app:algorithms}), these traces suggest that effective compression is associated with controlled length reduction while maintaining comparatively stable entropy and reward trajectories, rather than applying an indiscriminately stronger length penalty.

\end{document}